\PassOptionsToPackage{table}{xcolor}

\documentclass[10pt, a4paper, logo]{googledeepmind}

\usepackage{times}
\usepackage{url}
\usepackage{booktabs}
\usepackage{graphicx}
\usepackage{subcaption}
\usepackage{array}
\usepackage{longtable}
\usepackage{nicefrac}
\usepackage{multirow}
\usepackage{float}
\usepackage{wrapfig}
\usepackage{stfloats}
\usepackage{textcomp}
\usepackage{verbatim}
\usepackage{enumitem}
\usepackage{siunitx}
\usepackage{colortbl}
\usepackage[normalem]{ulem}
\usepackage{etoolbox}
\usepackage[most]{tcolorbox}
\tcbuselibrary{breakable}
\usepackage[ruled,linesnumbered,lined]{algorithm2e}
\usepackage[leftcaption]{sidecap}
\usepackage{natbib}

\usepackage{amsmath,amsfonts,bm}

\def\eqref#1{equation~\ref{#1}}

\def\1{\bm{1}}

\DeclareMathAlphabet{\mathsfit}{\encodingdefault}{\sfdefault}{m}{sl}
\SetMathAlphabet{\mathsfit}{bold}{\encodingdefault}{\sfdefault}{bx}{n}

\definecolor{BestColor}{HTML}{C8E6C9}
\definecolor{SecondBestColor}{HTML}{FFF9C4}
\definecolor{impcolor}{HTML}{2E8B57}
\definecolor{ggg}{RGB}{26,179,0}
\definecolor{rrr}{RGB}{179,0,0}
\definecolor{oodc}{RGB}{31,73,121}
\definecolor{idc}{RGB}{68,142,68}
\definecolor{mygray}{gray}{0.9}

\newcommand{\best}[1]{\cellcolor{BestColor}\textbf{#1}}
\newcommand{\bestmath}[1]{\cellcolor{BestColor}{\boldmath\textbf{#1}}}
\newcommand{\secondbest}[1]{\cellcolor{SecondBestColor}#1} \newcommand{\gainhl}[1]{{\setlength{\fboxsep}{1pt}\colorbox{BestColor}{\textbf{#1}}}} \DeclareRobustCommand{\capbest}[1]{{\setlength{\fboxsep}{1pt}\colorbox{BestColor}{\textbf{#1}}}} \DeclareRobustCommand{\capsecond}[1]{{\setlength{\fboxsep}{1pt}\colorbox{SecondBestColor}{#1}}}
\newcommand{\improvementstyle}[1]{$^{\textcolor{impcolor}{\tiny #1}}$}
\newcommand{\scoreimp}[2]{%
  \textbf{#1}%
  \ifstrequal{#2}{+0.0}{}{%
    \ifstrequal{#2}{0.0}{}{%
      \makebox[0pt][l]{\improvementstyle{#2}}%
    }%
  }%
}

\def\Bias#1#2{\bm{b}}

\newtcolorbox{examplebox}[2][]{
    breakable,
    enhanced,
    colback=white,
    colframe=cyan,
    coltitle=white,
    fonttitle=\bfseries,
    title=#2,
    overlay middle={\draw[cyan, line width=1pt](frame.south west)--(frame.south east);},
    overlay last={\draw[cyan, line width=1pt](frame.south west)--(frame.south east);},
    #1
}

\newenvironment{nolinenumbers}{}{}

\newenvironment{responsebox}[1]{%
    \begin{nolinenumbers}
    \begin{tcolorbox}[
        breakable,
        enhanced,
        colback=gray!4,
        colframe=gray!45,
        colbacktitle=gray!45,
        coltitle=white,
        fonttitle=\bfseries,
        title={#1},
        boxrule=0.5pt,
        arc=1.5mm,
        left=2mm,
        right=2mm,
        top=1.5mm,
        bottom=1.5mm,
        before skip=0.7em,
        after skip=0.7em
    ]
}{%
    \end{tcolorbox}
    \end{nolinenumbers}
    \normalsize
}

\newenvironment{takeawaybox}{%
    \begin{tcolorbox}[
        enhanced,
        colback=gray!6,
        colframe=gray!35,
        boxrule=0.4pt,
        arc=1mm,
        left=1.5mm,
        right=1.5mm,
        top=0.8mm,
        bottom=0.8mm,
        before skip=0.45em,
        after skip=0.45em
    ]
}{%
    \end{tcolorbox}
}

\newcommand{\gapmethod}{GAP (LH+PCA+DA)}

\title{Fill the GAP: A Granular Alignment Paradigm for Visual Reasoning in Multimodal Large Language Models}

\author[1,2,4]{Yanting Miao\textsuperscript{*}}
\author[1,3]{Yutao Sun\textsuperscript{*}}
\author[1]{Dexin Wang}
\author[1]{Mengyu Zhou\textsuperscript{$\dagger$}}
\author[2,4]{Pascal Poupart}
\author[1]{Lei Lv}
\author[1]{Qi Zhao}
\author[1]{Li Wang}
\author[1]{Hao Li}
\author[1]{Xiaoxi Jiang}
\author[1]{Guanjun Jiang}

\affil[1]{Qwen Business Unit, Alibaba}
\affil[2]{University of Waterloo}
\affil[3]{Zhejiang University}
\affil[4]{Vector Institute}

\correspondingauthor{Mengyu Zhou (\texttt{zhoumengyu.zmy@alibaba-inc.com})}

\begin{abstract}
Visual latent reasoning lets a multimodal large language model (MLLM) create intermediate visual evidence as continuous tokens, avoiding external tools or image generators.
However, existing methods usually follow an output-as-input latent paradigm and yield unstable gains.
We identify a feature-space mismatch behind this instability: dominant visual-latent models build on pre-norm MLLMs and reuse decoder hidden states as predicted latent inputs, even though these states occupy a substantially different norm regime from the input embeddings the model was trained to consume~\citep{xie2025mhc,li2026siamesenorm,team2026attention}.
This mismatch makes direct latent feedback unreliable.
Motivated by this diagnosis, we propose \textbf{GAP}, a \textbf{G}ranular \textbf{A}lignment \textbf{P}aradigm for visual latent modeling.
GAP aligns visual latent reasoning at three levels: data-level alignment grounds latent targets with inspectable auxiliary visual supervision; feature-level alignment maps decoder outputs into input-compatible visual latents through a lightweight PCA-aligned latent head; and model-level alignment assigns latent supervision selectively to examples where the base MLLM struggles.
On Qwen2.5-VL 7B, GAP achieves the best aggregate perception and reasoning scores among all compared methods and is the only evaluated latent method that improves both over the base model.
Inference-time interventions further show that generated latents carry task-relevant visual signal rather than merely adding token slots.
Code is available at \url{https://github.com/Qwen-Applications/GAP}.
\end{abstract}

\begin{document}
\maketitle

\section{Introduction}
\label{sec:introduction}

Multimodal large language models (MLLMs) increasingly solve visual tasks by interleaving perception and language reasoning, yet many failures on charts, high-resolution perception, and visual math stem from missing or poorly localized visual evidence ~\citep{bai2025qwen3vltechnicalreport,singh2025openai,team2025gemma}.
A natural way to obtain such evidence is to call external visual tools, such as cropping, search, or image generation~\citep{miao2024subject}, to create auxiliary views.
Tool-augmented pipelines, however, introduce extra API calls, wall-clock latency, and system complexity, and building a unified generation-and-understanding model is a substantially heavier direction that typically requires many additional visual tokens and training resources.
Tool-based approaches also face cross-task generalization difficulties, since a single visual tool is rarely adequate across diverse visual reasoning tasks~\citep{su2025thinking, zhao2025pyvision}.
We instead ask whether an image-understanding MLLM can be minimally adapted to produce intermediate visual evidence inside its own autoregressive computation.
One appealing solution is \emph{visual latent reasoning}: the model emits continuous latent tokens intended to represent auxiliary visual information~\citep{wang2025monet,li2025latent}.
This design is lightweight and keeps inference inside a single model, with no external image-generation call.
However, prior latent-token methods deliver unstable empirical gains, and in some settings the latent channel contributes little after post-training~\citep{li2025imagine,wu2026lavit, liu2025reasoning}.

\textbf{A feature-level failure mode in pre-norm MLLMs.}
Modern MLLMs such as Qwen3-VL~\citep{bai2025qwen3vltechnicalreport}, Gemma 4~\citep{googledeepmind2026gemma4modelcard}, and InternVL3.5~\citep{wang2025internvl3} use pre-norm transformers, in which final-layer hidden states accumulate much larger norms than the input text or vision embeddings and are not guaranteed to lie in the same empirical subspace as input vision embeddings.
Most recent visual-latent models build on pre-norm MLLM backbones and reuse output hidden states as next-step latent inputs~\citep{wang2025monet,li2025latent}, implicitly assuming that these outputs are valid inputs.
We test this assumption on Monet-7B~\citep{wang2025monet}, a state-of-the-art latent model based on Qwen2.5-VL-7B and find that the state it feeds back as the next latent is still about $4.7\times$ larger in norm than the input vision embeddings, even after the backbone's final RMSNorm (Figure~\ref{fig:norm_plot}(b)).
Re-injecting such states creates a feature-space mismatched feedback loop, in which every latent step drives the model with inputs from a distribution it was not trained to consume.

\textbf{Interventional evidence.}
The norm gap is not only descriptive.
Taking Monet-7B as an output-as-input latent baseline and applying a training-free Exponential Moving Average (EMA) rescaling that matches predicted latent norms to vision input-embedding norms improves performance by $+2.00$ on MathVista.
The intervention changes neither the training data nor the backbone; it only corrects the norm of the re-injected latent vector.
This supports feature-level compatibility as a practical factor for stable latent reuse. Even without changing data or backbone, correcting the re-injected latent's magnitude already recovers measurable accuracy, and the residual gap motivates the learned alignment we introduce next.

\begin{figure}[t]
\centering
\includegraphics[width=\linewidth]{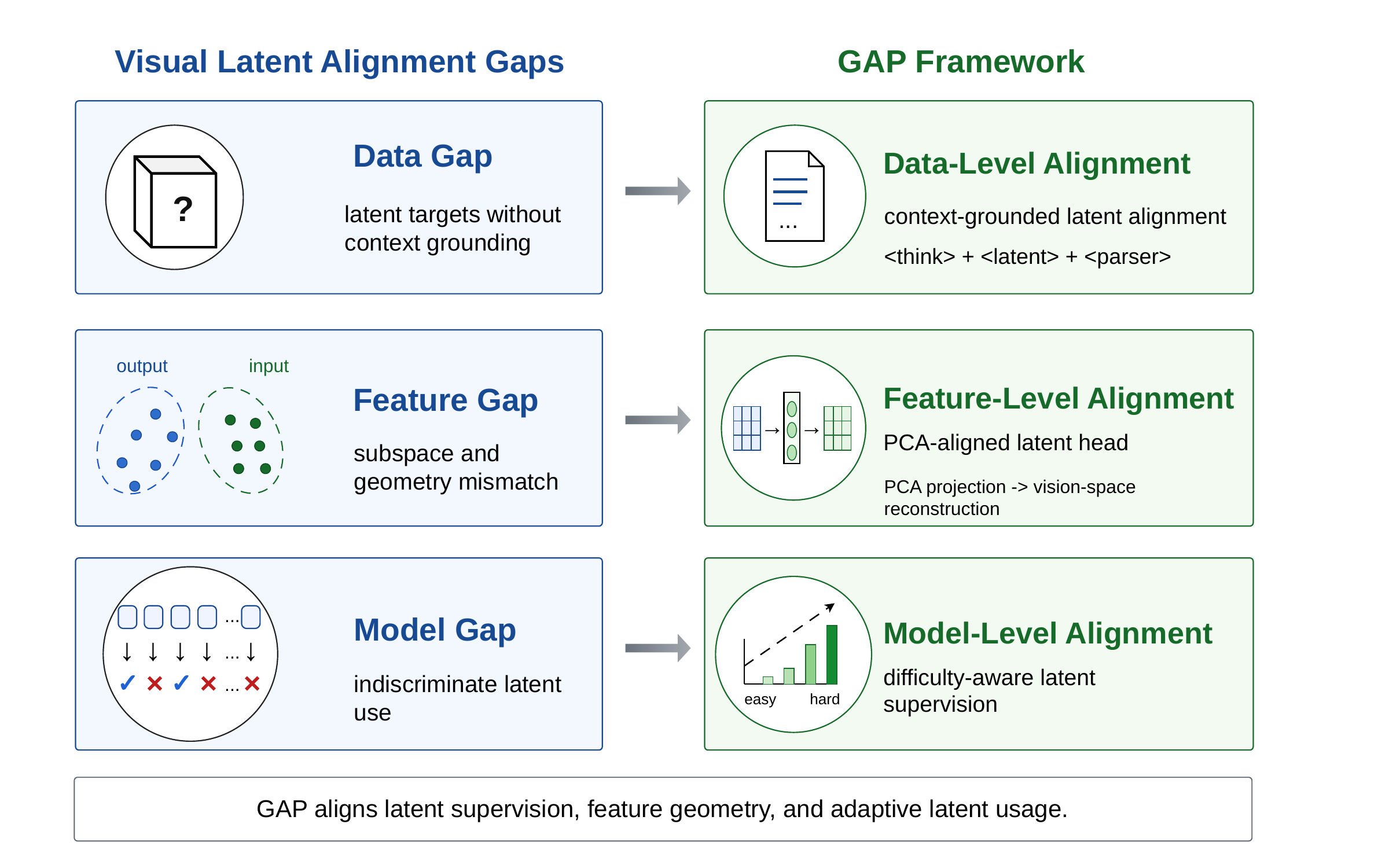}
\caption{Conceptual overview of GAP.
GAP addresses visual latent learning through three alignment levels: context-grounded latent supervision at the data level, a PCA-aligned latent head at the feature level, and difficulty-aware latent supervision at the model level.}
\label{fig:gap_conceptual}
\end{figure}

\textbf{Our approach: GAP.}
The feature-level mismatch above is the main failure mode targeted in this work, and norm rescaling alone treats only one symptom of it.
Training a useful latent channel also benefits from two practical supports: the latent targets should be interpretable enough to inspect during training and analysis, and latent supervision should not be applied indiscriminately to examples the base model already solves.
We therefore introduce \textbf{GAP}, a \textbf{G}ranular \textbf{A}lignment \textbf{P}aradigm for visual latent modeling across data, feature, and model levels.
At the feature level, a \textbf{PCA-aligned latent head} maps the backbone's native-RMS-normalized decoder states into the principal subspace of auxiliary-image vision embeddings before re-injection; the PCA basis acts both as a subspace projector that returns generated latents to the empirical vision-embedding distribution and as a low-rank parameterization of the latent head.
For inspectability, \textbf{context-grounded latent supervision} annotates each latent example with chain-of-thought intent, a \texttt{<latent>} span supervised by auxiliary-image vision embeddings, and a \texttt{<parser>} description that records the intended auxiliary visual signal during training and analysis.
For selective supervision, \textbf{difficulty-aware latent assignment} estimates base-model accuracy by repeated sampling and assigns latent targets only to examples the base model fails to solve, while retaining text supervision on easier examples.
Figure~\ref{fig:gap_conceptual} summarizes GAP as a three-level alignment recipe spanning data-, feature-, and model-level alignment.
The resulting training set contains 49K curated multimodal examples spanning visual CoT, charts, geometry, multimodal math, counting, and visual search.

We make the following contributions:
\begin{enumerate}[leftmargin=*]
    \item \textbf{A granular alignment paradigm (GAP).} We introduce a lightweight latent feedback recipe centered on a PCA-aligned projection that returns generated latents to the vision-embedding subspace, supported by context-grounded latent supervision for inspectability and difficulty-aware latent assignment for selective supervision.
    \item \textbf{Empirical validation of aligned visual latents.}
        Across five perception and reasoning benchmarks with a Qwen2.5-VL-7B backbone, GAP achieves
        the best aggregate perception and reasoning scores over prior visual-latent baselines and our
        supervised variants.
        Controlled ablations show that PCA alignment, curated latent supervision, and
        difficulty-aware assignment each contribute to the gains, and inference-time
        interventions show that generated latents carry task-relevant visual signal
        beyond merely adding token slots.
    \item \textbf{Curated latent-supervision data.} We construct a 49K high-quality multimodal latent-supervision dataset spanning visual CoT, charts, geometry, multimodal math, counting, and visual search. We release it through our code repository to support future work on visual latent reasoning.
\end{enumerate}

Beyond the high-level overview in Figure~\ref{fig:gap_conceptual}, Appendix Figure~\ref{fig:architecture} traces the token-level execution path, showing how ordinary text decoding and GAP latent reconstruction interleave during generation.

\section{Related Work}
\label{sec:related}

\textbf{Explicit visual tools and generated visual thoughts.}
One way to supply missing visual evidence is to call external visual operations such as zooming, cropping, or region search.
ZoomEye~\citep{shen2025zoomeye} builds a tree-based exploration strategy that enables MLLMs to inspect relevant image regions.
Another line generates explicit visual thoughts or auxiliary images: Visualization-of-Thought~\citep{li2025imagine}, Latent Sketchpad~\citep{zhang2025latent}, Render-of-Thought~\citep{wang2026render}, DeepEyes~\citep{zheng2025deepeyes}, and GoT-R1~\citep{duan2025got} use sketching, rendered reasoning traces, visual tools, or generation-oriented reinforcement learning.
These methods provide interpretable intermediate artifacts, but they introduce additional tool or generation dependencies at inference.
Our narrower goal is to keep inference inside the MLLM and to study how continuous visual feedback can be made compatible with the model's input-embedding distribution.

\textbf{Continuous and visual latent reasoning.}
Latent reasoning represents intermediate computation as continuous embeddings rather than explicit text or images.
\citet{deng2024explicit} study the transition from explicit to implicit chain-of-thought, and COCONUT~\citep{hao2024training} trains language models to reason in a continuous latent space.
In multimodal reasoning, Monet~\citep{wang2025monet}, LVR~\citep{li2025latent}, Machine Mental Imagery~\citep{yang2025machine}, Chain-of-Visual-Thought~\citep{qin2025chain}, LIVR~\citep{li2025latentimplicit}, VaLR~\citep{jeon2026vision}, LaViT~\citep{wu2026lavit}, and CrystaL~\citep{zhang2026crystal} explore latent visual tokens, vision-aligned latent thoughts, or emergent visual latents.
These works make visual latent reasoning an active and competitive direction; our contribution is therefore not the introduction of visual latent tokens themselves.
Instead, we isolate a concrete feature-space failure mode of output-as-input latent feedback in pre-norm MLLMs and test a lightweight PCA-based reconstruction path that maps generated latents back toward the empirical vision-embedding subspace before autoregressive re-injection.

\textbf{Residual norms and input compatibility in pre-norm transformers.}
Pre-norm transformers stabilize deep training by normalizing each sublayer input, but their residual streams can accumulate large norms across depth.
Recent analyses and architectures such as SiameseNorm~\citep{li2026siamesenorm}, Attention Residuals~\citep{team2026attention}, and manifold-constrained hyper-connections~\citep{xie2025mhc} study the behavior of residual streams and normalization.
Our work connects this architectural issue to visual latent feedback. A vector that is suitable for output readout need not be distributed like an input vision embedding, even after the backbone's native final RMSNorm.

\textbf{Selective latent computation.}
Latent and visual computation need not be applied uniformly to every example.
Visual Enhanced Depth Scaling~\citep{han2026visual} studies when additional visual computation should be allocated during multimodal latent reasoning.
Our difficulty-aware supervision is a training-time counterpart that decides which examples should include latent targets, without adding an inference-time routing or depth-scaling module.

\section{Preliminary Analysis: Why Output-as-Input Latent Feedback Fails}
\label{sec:analysis}

Before presenting our method, we first establish the feature-space bottleneck behind output-as-input latent feedback in our Qwen2.5-VL-7B setting.
Section~\ref{sec:component_ablation} tests whether the latent head, curated data, difficulty-aware supervision, and PCA compression each matter, using component ablations under controlled variants.

\subsection{Output-Embedding Norm Mismatch}
\label{sec:norm_mismatch}

Many current MLLM decoders, including Qwen2.5-VL, adopt a \emph{pre-norm} architecture~\citep{xiong2020layer}, where layer normalization is applied \emph{before} each sub-layer (attention and FFN) rather than after.
In pre-norm decoders, residual updates are added to an unnormalized stream, allowing hidden-state norms to accumulate with depth.

\begin{figure}[t]
\centering
\begin{subfigure}[t]{0.49\linewidth}
    \centering
    \includegraphics[width=\linewidth]{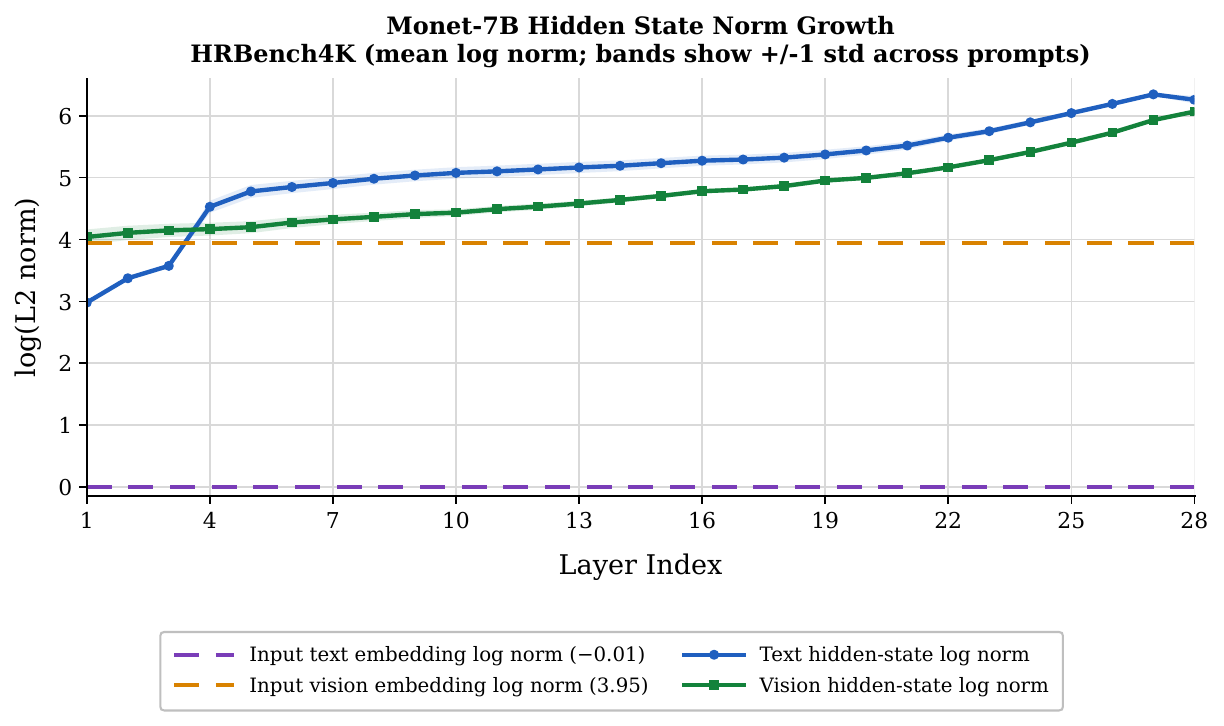}
    \caption{Layer-wise hidden-state norm growth.}
\end{subfigure}
\hfill
\begin{subfigure}[t]{0.49\linewidth}
    \centering
    \includegraphics[width=\linewidth]{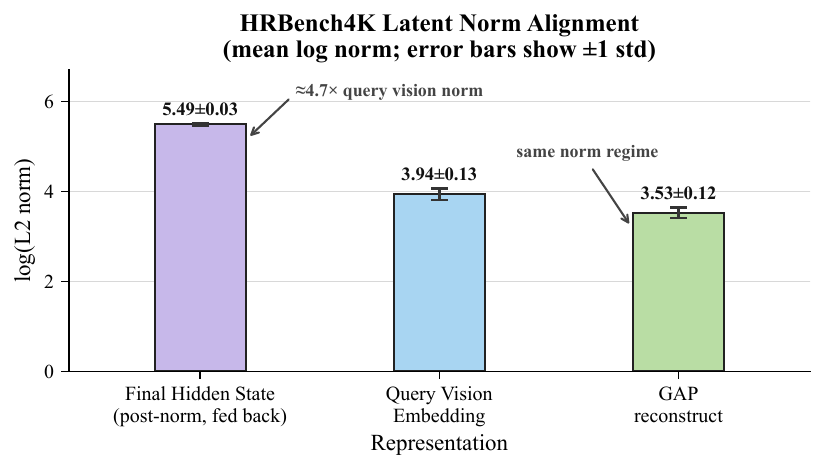}
    \caption{Latent norm alignment after GAP reconstruction.}
\end{subfigure}
\caption{Feature-space mismatch and GAP reconstruction on HRBench4K samples. (a) In Monet-7B, text and vision hidden-state log norms, measured before the final RMSNorm, grow sharply through the decoder stack and remain far above the input embedding regime; shaded bands show one standard deviation across prompts of $\log(\mathrm{L2\ norm})$ for hidden-state norms only. (b) The final decoder hidden state that output-as-input latent feedback reuses, taken after the final RMSNorm, is still about $4.7\times$ larger than the query vision-embedding norm, whereas GAP reconstruction maps generated latents back to the same norm regime before autoregressive re-injection.}
\label{fig:norm_plot}
\end{figure}

We profile the hidden-state norms of Monet-7B on HRBench4K samples (Figure~\ref{fig:norm_plot}).
Monet feeds back the decoder state after the final RMSNorm as the next latent input.
This post-norm state is still about $4.7\times$ the input vision-embedding norm at vision positions (Figure~\ref{fig:norm_plot}(b)) and about $290\times$ the input text-embedding norm at text positions.
The final RMSNorm first normalizes each state to approximately unit RMS and then applies the learned elementwise gain $\gamma$ ($\|\gamma\|_2 = 240.7$, much larger than the input-embedding norms).
Before this norm, text and vision hidden-state norms grow steadily through the decoder stack (Figure~\ref{fig:norm_plot}(a)).
The growth persists after Monet's latent fine-tuning, so Monet inherits it from its pre-norm Qwen2.5-VL-7B decoder.
Let $\mathbf{h}^{(l)} \in \mathbb{R}^d$ denote the hidden state at layer $l$ ($d = 3584$).
The growth reflects residual accumulation in pre-norm architectures, where each transformer layer computes $\mathbf{h}^{(l+1)} = \mathbf{h}^{(l)} + \text{SubLayer}(\text{LN}(\mathbf{h}^{(l)}))$, where $\text{LN}$ denotes layer normalization, $\text{SubLayer}$ abstracts the layer's attention or feed-forward transformation, and the additive residual can accumulate norm across layers~\citep{li2026siamesenorm,xie2025mhc,team2026attention}.

Existing latent reasoning methods~\citep{hao2024training, wang2025monet,li2025latent} directly reuse the output hidden state $\mathbf{h}^{(L)}$ as the input embedding for the next latent token.
This feeds a vector from a high-norm output-side distribution into a processing pipeline calibrated for the input-embedding norm regime, creating a distribution shift that destabilizes optimization.
Thus, direct output-as-input reuse creates a feature-space mismatch in the latent feedback loop.
For comparison, Figure~\ref{fig:norm_plot}(b) shows that GAP's reconstruction path returns latents to the query vision-embedding norm regime, while the post-norm final decoder hidden state that output-as-input reuse feeds back remains about $4.7\times$ larger.

\textbf{Empirical evidence: norm rescaling alone helps.}
To validate that the norm gap is not merely a theoretical concern but a practical bottleneck, we conduct a simple experiment.
We use Monet-7B as an output-as-input latent model, and compare it with and without a training-free EMA norm calibration that rescales the predicted latent to match the input vision embedding norm:
$\tilde{\mathbf{v}}_t = \bar{n}_{\text{EMA}} \cdot {\hat{\mathbf{v}}_t}/{\|\hat{\mathbf{v}}_t\|_2}$,
where $\bar{n}_{\text{EMA}}$ is the running average of $\|\mathbf{v}\|_2$ computed from the query image's vision embedding.
As shown in Table~\ref{tab:norm_mismatch}, simply rescaling the norm at inference time already improves performance, confirming norm mismatch as a practical bottleneck.

\begin{table}[H]
\centering
\caption{Effect of norm calibration. ``Output-as-Input'' denotes Monet-7B, a latent-reasoning model trained from Qwen2.5-VL-7B that reuses output-side hidden states as latent inputs. EMA norm calibration rescales the predicted latent to match input vision embedding norms at inference time, requiring no additional training. The improvement row is highlighted in green.}
\label{tab:norm_mismatch}
\begin{tabular}{l ccc}
\toprule
\textbf{Configuration} & \textbf{HRBench4K}$\uparrow$ & \textbf{MathVista}$\uparrow$ & \textbf{Avg.}$\uparrow$ \\
\midrule
Output-as-Input (Monet-7B) & 70.75 & 61.30 & 66.03 \\
Output-as-Input + EMA norm calibration & 71.63 & 63.30 & 67.46 \\
\midrule
\textit{Improvement} & \cellcolor{BestColor}\textit{+0.88} & \cellcolor{BestColor}\textit{+2.00} & \cellcolor{BestColor}\textit{+1.44} \\
\bottomrule
\end{tabular}
\end{table}

The norm intervention motivates GAP's feature-level design, in which the model does not recycle raw decoder outputs as future visual inputs.
We next describe the data-, feature-, and model-level alignment mechanisms, and then report controlled mechanism, data, PCA-rank, and token-budget ablations in Section~\ref{sec:experiments}.

\section{Method}
\label{sec:method}

Guided by the alignment requirements in Section~\ref{sec:introduction}, GAP builds latent reasoning around three levels of alignment.
Data-level alignment makes each continuous target traceable to a controllable auxiliary visual signal (\S\ref{sec:data}); feature-level alignment maps generated latents back into the empirical vision-input subspace before feedback (\S\ref{sec:latent_head}); and model-level alignment applies latent supervision mainly to queries where the base model struggles (\S\ref{sec:difficulty_aware}).

\subsection{Data-Level Alignment: Controllable Latent Data Construction}
\label{sec:data}

Continuous latent supervision is otherwise hard to inspect. A latent vector can be supervised, but the example alone does not reveal what missing visual evidence that vector is meant to represent.
We therefore construct 49,309 multimodal QA examples in which each query image and question is paired with an intermediate auxiliary image and a structured teacher response.
During student training, the auxiliary image is not supplied as an image input to the student. Its frozen-ViT features fill the teacher-forced latent positions and also serve as targets for the latent head, while the surrounding \texttt{<think>} and \texttt{<parser>} text records the intended visual content.
At inference time no auxiliary image is available, and the same latent positions are predicted autoregressively by GAP.
The dataset is concentrated on visual CoT, chart understanding, GEOQA, multimodal math, counting, and visual search; full source composition and exact latent serialization are provided in Appendix~\ref{app:data}, with data-loader details in Appendix~\ref{app:impl}.
Auxiliary images are generated with Nano Banana 2 and structured responses are synthesized offline by Qwen3-VL-235B-A22B-Thinking; we estimate the construction cost at approximately US\$3.31K in image-generation API fees and 274 GPU-hours for teacher synthesis (Appendix~\ref{app:latency}).

\begin{tcolorbox}[
    colback=gray!6,
    colframe=gray!35,
    boxrule=0.4pt,
    arc=1mm,
    left=1.5mm,
    right=1.5mm,
    top=0.8mm,
    bottom=0.8mm
]
\small
\textbf{Latent Interleave response format.}
Text reasoning $\rightarrow$ \texttt{<latent>} visual latent tokens \texttt{</latent>}
$\rightarrow$ \texttt{<parser>} textual description of the intended latent visual evidence \texttt{</parser>}
$\rightarrow$ continued text reasoning and final answer.
\end{tcolorbox}

\subsection{Feature-Level Alignment: PCA-Aligned Latent Head}
\label{sec:latent_head}

At the feature level, the decoder state used to generate a latent token need not be distributed like the vision embeddings consumed as input.
GAP avoids direct output-as-input reuse by predicting low-dimensional coordinates in an empirical auxiliary-image vision subspace and reconstructing each generated latent in the input vision-embedding coordinate system before re-injection.
Let $\hat{\mathbf{v}}_{t-1}\in\mathbb{R}^d$ be the previous latent embedding fed to the transformer, $\mathbf{h}_t^{(L)}$ is the hidden state at the layer $L$, $\mathcal{M}_{\theta}$ the backbone, and $\mathcal{F}_{\theta}$ the latent head.
Autoregressive latent generation is:
\begin{equation}
    \mathbf{h}_t^{(L)}=\mathcal{M}_{\theta}(\hat{\mathbf{v}}_{t-1},\mathrm{KV}_{<t}),\quad \mathbf{c}_t=\mathcal{F}_{\theta}(\bar{\mathbf{h}}_t),\quad \hat{\mathbf{v}}_t=\mathbf{P}_k\mathbf{c}_t+\boldsymbol{\mu}.
    \label{eq:latent_generation}
\end{equation}
where $\bar{\mathbf{h}}_t = \text{RMSNorm}(\mathbf{h}_t^{(L)})$, $\mathrm{KV}_{<t}$ denotes cached keys and values, and $\mathbf{c}_t\in\mathbb{R}^k$ are PCA coefficients.
We estimate $\mathbf{P}_k$ and $\boldsymbol{\mu}$ once from training-only auxiliary-image embeddings; validation and benchmark embeddings are excluded.
The basis is encoder-specific and frozen, and is reused across checkpoints, seeds, and tasks sharing the same encoder and fitting data.
All experiments use 95\% retained variance as a shared default, and Appendix~\ref{app:rebuttal_controls} examines how the selected rank depends on the fitting distribution.
PCA constrains the auxiliary latent pathway, while the original-image tokens remain accessible to the backbone.
A full $\mathbb{R}^d\!\to\!\mathbb{R}^d$ latent MLP would require roughly $2d^2 \approx 25.7$M parameters for $d=3584$; GAP instead uses $k=629$ components, retaining 95\% variance, and supervises the decoded embedding:
\begin{equation}
    \mathcal{L}_{\text{latent}} = \frac{1}{\lvert \mathcal{T} \rvert}
    \sum_{t \in \mathcal{T}} \left\|
    \mathbf{P}_k\mathcal{F}_\theta(\bar{\mathbf{h}}_t) + \boldsymbol{\mu} - \mathbf{v}_t
    \right\|_2^2,
    \label{eq:latent_loss}
\end{equation}
where $\mathcal{T}$ is the set of latent token positions and $\mathbf{v}_t$ is the target auxiliary-image embedding.
This reconstruction path is the feature-level alignment mechanism. Raw decoder states are never fed back directly, and Figure~\ref{fig:norm_plot}(b) shows that GAP reconstruction returns generated latents to the query vision-embedding norm regime.
The full objective combines language modeling with latent alignment:
\begin{equation}
    \mathcal{L} = \mathcal{L}_{\text{LM}} + \lambda_{\text{latent}} \mathcal{L}_{\text{latent}},
    \label{eq:total_loss}
\end{equation}
Appendix Figure~\ref{fig:architecture} illustrates the text-token versus latent-token inference path, and Appendix~\ref{app:impl} gives the deferred implementation details, including projector initialization and scheduled sampling ablations.

\subsection{Model-Level Alignment: Difficulty-Aware Latent Supervision}
\label{sec:difficulty_aware}

Not all queries benefit from latent reasoning. The matched comparison in Section~\ref{sec:component_ablation} shows that restricting latent supervision to examples the base model cannot solve improves all seven reported metrics.
For each training query, we run $N=8$ inference passes with the base Qwen2.5-VL 7B model, compute empirical accuracy $\hat{a}$, and assign supervision by:
\begin{equation}
    \text{DA}(q) = \begin{cases}
        \text{text-only} & \text{if } \hat{a} > \tau, \\
        \text{latent} & \text{if } \hat{a} \leq \tau,
    \end{cases}
    \label{eq:difficulty_aware}
\end{equation}
In the current experiments $\tau=0$, so only examples the base model never solves across the eight samples receive latent supervision.
For ``text-only'' queries, the model is trained with a standard text response; for ``latent'' queries, it uses the full latent format from Section~\ref{sec:data}.
This model-level alignment teaches when latent-assisted reasoning is useful without adding an inference-time router, and the exact data-loader stripping rule is deferred to Appendix~\ref{app:impl}.

\section{Experiments}
\label{sec:experiments}

\subsection{Experimental Setup}
\label{sec:setup}

\textbf{Base model and baselines.}
We use Qwen2.5-VL 7B~\citep{bai2025qwen25vltechnicalreport} as the base MLLM.
We compare against the zero-shot base model, a \textbf{Dense Caption SFT} control trained on the same 49K curated data without latent tokens or auxiliary-image alignment, Monet~\citep{wang2025monet}, and LVR~\citep{li2025latent}.
Our variants are a basic latent-head SFT model (LH), a difficulty-aware full latent head without PCA (LH+DA), an all-latent PCA-aligned latent-head model (LH+PCA), and the full difficulty-aware PCA-aligned version, \gapmethod.
Monet and the original LVR are prior-system baselines with their own training recipes and data. We additionally retrain LVR on the same 49K corpus (LVR-49K) to control the backbone and data; matched component ablations then examine PCA and difficulty-aware assignment within our recipe.

\textbf{Benchmarks and metrics.}
Following the evaluation of Monet and LVR, we evaluate perception on HRBench4K, MMStar, and MME-RealWorld-Lite, and additionally evaluate reasoning on MathVista and WeMath.
For HRBench4K, we report Overall, fine-grained spatial perception (FSP), and fine-grained counting perception (FCP).
For MME-RealWorld-Lite, we report Overall, Reasoning, and Perception sub-scores.
We define Avg-P as the average of HRBench4K Overall, MMStar, and MME-RealWorld-Lite Overall, and Avg-R as the average of MathVista, WeMath-S, and WeMath-L.

\textbf{Implementation details.}
We fine-tune Qwen2.5-VL 7B using AdamW with a base learning rate of $1{\times}10^{-5}$, a latent-head learning rate of $1{\times}10^{-5}$, a global batch size of 128, and 2 epochs on the 33,279-example training split of the 49K curated dataset. Dense Caption SFT, LVR-49K, and all GAP variants use this same training split.
The PCA basis is computed offline from training-set vision embeddings with a 95\% explained-variance target, yielding $k = 629$ principal components.
The latent loss weight is $\lambda_{\text{latent}} = 1.0$.
The main difficulty-aware run uses 36 latent tokens; Appendix~\ref{sec:token_budget_cost} reports the token-budget sweep, and Appendix~\ref{app:latency} reports wall-clock inference cost.
We use one shared budget of 36 tokens across benchmarks because it gives the best mean over HRBench4K, MMStar, and MathVista in our sweep, although 16 tokens is close and can be more efficient. Token count controls a perception--reasoning and accuracy--cost trade-off; the 144-token MME analysis in Appendix~\ref{app:rebuttal_controls} is a separate capacity comparison, not a benchmark-specific replacement of the default model.
Appendix Table~\ref{tab:latent_budget_seed_std} reports three-evaluator-seed mean $\pm$ standard deviation scores for each token budget.

\subsection{Main Results}
\label{sec:main_results}

Tables~\ref{tab:main_perception} and~\ref{tab:main_reasoning} evaluate the overall effectiveness of GAP.
We first ask whether the three-level alignment recipe improves both perception and reasoning.

\textbf{Comparison with prior latent methods.}
\gapmethod\ denotes the full method with a latent head (LH), PCA-aligned latent reconstruction (PCA), and difficulty-aware latent assignment (DA).
The prior latent baselines improve perception but lose reasoning accuracy relative to the zero-shot Qwen2.5-VL 7B baseline: Avg-R drops from 52.62 to 47.99 for Monet and 47.67 for LVR, matching the norm-mismatch failure mode analyzed in Section~\ref{sec:norm_mismatch}.
Monet and LVR use their own training recipes and data, so these rows are prior-system comparisons; the LVR-49K control below matches the data.
Under this comparison, Ours-\gapmethod\ improves Avg-P by \gainhl{+1.89} and Avg-R by \gainhl{+5.98} over Monet, and improves Avg-P by \gainhl{+0.71} and Avg-R by \gainhl{+6.30} over LVR.

\textbf{Separating data and method effects.}
Retraining LVR on the 49K corpus improves Avg-P from 60.76 to 60.86 and Avg-R from 47.67 to 49.69, confirming a contribution from the curated data.
With the backbone and data fixed, GAP further improves these aggregates by +0.61 and +4.28 over LVR-49K.
This comparison separates the data change from the remaining method differences; Appendix Table~\ref{tab:attribution} summarizes it together with the matched component comparisons that isolate latent supervision, PCA alignment, and DA.
Gains on individual benchmarks are not uniform. On MME-RW-Lite, LVR-49K and the default GAP both trail the original LVR on Overall; we attribute this to the limited coverage of real-world scenarios in our training mixture, and a larger latent budget recovers most of the gap (Appendix~\ref{app:mme_budget}).

\begin{table}[!ht]
\centering
\caption{Main perception results. HRBench4K reports Overall, FSP, and FCP; MMStar reports Overall; MME-RealWorld-Lite reports Overall, Reasoning, and Perception. Avg-P averages HRBench4K Overall, MMStar Overall, and MME-RealWorld-Lite Overall. Best results are in \capbest{green} and second-best in \capsecond{yellow}.}
\label{tab:main_perception}
\begingroup
\setlength{\tabcolsep}{3pt}
\begin{tabular}{@{}l ccc c ccc c@{}}
\toprule
\multirow{2}{*}{\textbf{Method}} & \multicolumn{3}{c}{\textbf{HRBench4K}} & \multicolumn{1}{c}{\textbf{MMStar}} & \multicolumn{3}{c}{\textbf{MME-RW-Lite}} & \multirow{2}{*}{\textbf{Avg-P}} \\
\cmidrule(lr){2-4} \cmidrule(lr){5-5} \cmidrule(lr){6-8}
& Ovr.$\uparrow$ & FSP$\uparrow$ & FCP$\uparrow$ & Ovr.$\uparrow$ & Ovr.$\uparrow$ & Rsn.$\uparrow$ & Perc.$\uparrow$ & \\
\midrule
Qwen2.5-VL & 68.13 & 83.50 & 52.75 & 60.46 & 44.39 & 37.60 & 48.76 & 57.66 \\
Dense Cap & 69.38 & 85.50 & 53.25 & 60.50 & 47.73 & 42.40 & 51.15 & 59.20 \\
Monet-7B & 70.75 & \best{90.50} & 51.00 & 60.80 & 47.20 & 42.90 & 50.00 & 59.58 \\
LVR & 70.75 & 84.25 & 57.25 & 62.07 & \best{49.45} & \secondbest{43.07} & \best{53.55} & 60.76 \\
LVR-49K & 71.63 & 84.50 & \secondbest{58.75} & \secondbest{62.17} & \secondbest{48.78} & \best{43.20} & \secondbest{52.36} & \secondbest{60.86} \\
\midrule
Ours-LH & 71.38 & 85.50 & 57.25 & 61.20 & 45.67 & 39.20 & 49.82 & 59.42 \\
Ours-LH+DA & 71.88 & 85.00 & \secondbest{58.75} & 61.80 & 45.75 & 39.06 & 50.00 & 59.81 \\
Ours-LH+PCA & \secondbest{72.25} & \secondbest{87.75} & 56.75 & 62.10 & 46.93 & 39.80 & 51.50 & 60.43 \\
Ours-\gapmethod & \best{73.25} & 86.50 & \best{60.00} & \best{63.40} & 47.77 & \secondbest{43.07} & 50.79 & \best{61.47} \\
\bottomrule
\end{tabular}
\endgroup
\end{table}

\begin{table}[!ht]
\centering
\caption{Main reasoning results. MathVista measures multimodal mathematical reasoning, and WeMath is reported with strict (S) and loose (L) accuracy. Avg-R averages MathVista, WeMath-S, and WeMath-L. Best results are in \capbest{green} and second-best in \capsecond{yellow}.}
\label{tab:main_reasoning}
\begin{tabular}{l ccc c}
\toprule
\textbf{Method} & \textbf{MathVista}$\uparrow$ & \textbf{WeMath-S}$\uparrow$ & \textbf{WeMath-L}$\uparrow$ & \textbf{Avg-R}$\uparrow$ \\
\midrule
Qwen2.5-VL 7B & 68.50 & \secondbest{36.29} & \secondbest{53.06} & \secondbest{52.62} \\
Dense Cap SFT & 67.70 & 28.60 & 45.43 & 47.24 \\
Monet-7B & 61.30 & 32.67 & 50.00 & 47.99 \\
LVR & 67.00 & 26.95 & 49.05 & 47.67 \\
LVR-49K & 68.82 & 30.59 & 49.65 & 49.69 \\
\midrule
Ours (LH) & 68.60 & 30.00 & 48.86 & 49.15 \\
Ours (LH+DA, no PCA) & 68.90 & 31.26 & 50.00 & 50.05 \\
Ours (LH+PCA) & \secondbest{70.20} & 35.24 & 52.00 & 52.48 \\
Ours-\gapmethod & \best{71.00} & \best{36.33} & \best{54.57} & \best{53.97} \\
\bottomrule
\end{tabular}
\end{table}

Ours-\gapmethod\ achieves the highest Avg-P and Avg-R of all methods and is the only latent method that improves both aggregates over the zero-shot base model.

\begin{takeawaybox}
\textbf{Takeaway 1.} The three-level alignment recipe gives the best aggregate perception and reasoning performance and is the only latent method that improves both over the base model.
\end{takeawaybox}

\textbf{Why dense captioning helps perception but hurts reasoning.}
Dense Caption SFT improves the perception aggregate from 57.66 to 59.20, because more detailed textual descriptions of the image help tasks that primarily require fine-grained visual recognition.
However, it still trails the base model on MathVista and substantially degrades WeMath, reducing the reasoning aggregate from 52.62 to 47.24.
We attribute this to a mismatch between dense descriptive supervision and multi-hop reasoning.

\textbf{Why visual latents need alignment and selectivity.}
The basic LH row improves over Dense Caption SFT on HRBench4K, MathVista, and both aggregates (+0.22 Avg-P, +1.91 Avg-R), but it still trails the base model on Avg-R (49.15 versus 52.62), so a latent head alone is not sufficient for robust aggregate gains.
PCA alignment addresses this by reconstructing generated latents in the empirical subspace spanned by real training-set vision embeddings, bringing the fed-back latents closer to the input vision-embedding subspace and norm regime.
It also reduces the latent-head dimensionality, lowering the risk that the full head overfits the curated latent-supervision distribution; on the OOD POPE check in Appendix~\ref{app:pope_ood}, adding PCA restores the overall score from 85.72 to 86.32, on par with the base model (86.23).
DA retains text supervision on examples the base model already solves and reserves latent targets for harder cases; the matched comparison in Section~\ref{sec:component_ablation} isolates its effect.
Neither partial variant exceeds the zero-shot base on Avg-R (50.05 for LH+DA and 52.48 for LH+PCA, versus 52.62), whereas combining both components reaches 53.97.
The two partial variants expose complementary failure modes rather than instability of the full recipe. LH+DA leaves the output--input feature mismatch unresolved, whereas LH+PCA imposes \emph{overly strong} latent supervision by forcing latent targets onto every example, including those the base model already solves.
The full method gains +1.35 Avg-R over the base model, and this gain requires both components.

\subsection{Component Ablations}
\label{sec:component_ablation}

We next separate the remaining ingredients behind GAP beyond the basic LH row in Tables~\ref{tab:main_perception} and~\ref{tab:main_reasoning}: curated latent data with difficulty-aware supervision and PCA-aligned latent generation.
These ablations complement the full benchmark suite above by isolating narrower two- or three-benchmark comparisons.

\textbf{Data quality and difficulty-aware supervision.}
Table~\ref{tab:data_effectiveness} compares latent-supervision variants that differ in data source and assignment strategy.
The curated 49K latent mixture improves substantially over training our latent-head recipe on Monet's 125K SFT data, especially on MathVista.
Without PCA, DA adds only +0.40 over all-latent supervision, since both variants still suffer from the unresolved feature mismatch. We therefore isolate DA inside the PCA-aligned recipe, comparing LH+PCA with LH+PCA+DA under the same training configuration and changing only the supervision assignment.

\begin{table}[!ht]
\centering
\caption{Component comparison for latent supervision data and assignment strategy. Best results are in \capbest{green} and second-best in \capsecond{yellow}.}
\label{tab:data_effectiveness}
\begin{tabular}{l c ccc}
\toprule
\textbf{Training Data / Strategy} & \textbf{Size} & \textbf{HRBench4K}$\uparrow$ & \textbf{MathVista}$\uparrow$ & \textbf{Avg.}$\uparrow$ \\
\midrule
Monet latent model & 125K & 70.75 & 61.30 & 66.03 \\
Ours (latent head on Monet SFT data) & 125K & 71.00 & 62.70 & 66.85 \\
Ours, all latent & 49K & \secondbest{71.38} & \secondbest{68.60} & \secondbest{69.99} \\
Ours, difficulty-aware (LH+DA) & 49K & \best{71.88} & \best{68.90} & \best{70.39} \\
\bottomrule
\end{tabular}
\end{table}

\begin{takeawaybox}
\textbf{Takeaway 2.} Within the PCA-aligned recipe, difficulty-aware assignment improves both aggregates in the main tables (+1.04 Avg-P, +1.49 Avg-R) and all seven metrics in a five-seed matched evaluation.
\end{takeawaybox}

\textbf{Matched difficulty-aware comparison.}
We evaluate fixed prediction sets from LH+PCA and LH+PCA+DA across five evaluator seeds (Appendix Table~\ref{tab:da_evaluator}).
DA increases HRBench4K Overall from $72.25\pm0.1581$ to $73.53\pm0.1559$ and MathVista from $70.06\pm0.2348$ to $71.20\pm0.1643$, and raises the mean on all seven reported metrics.
The two models differ only in which examples receive latent targets, so these gains come from withholding latent supervision on examples the base model already solves, which removes the overly strong supervision of LH+PCA.

\textbf{PCA components and capacity control.}
Table~\ref{tab:pca_rank_ablation_main} varies the number of retained PCA components while keeping the backbone, curated training data, 36 latent tokens, difficulty-aware supervision, and evaluation benchmarks fixed.
The uncompressed LH+DA head already improves over the base model, but all PCA+DA variants further improve Avg-3 while predicting only 7--18\% as many output dimensions, so the input-space PCA parameterization helps beyond head capacity.
The default 95\% setting gives the best three-benchmark mean; Appendix~\ref{app:rebuttal_controls} examines its sensitivity to the fitting distribution and shows that GAP improves the MathVista geometry subset, where fine geometric detail matters, by +5.39 points over the base model.

\begin{table}[!ht]
\centering
\caption{PCA component-count and capacity-control ablation under difficulty-aware latent supervision. Avg-3 averages HRBench4K Overall, MMStar, and MathVista. $\Delta$ is measured relative to Qwen2.5-VL-7B Avg-3. Best results are in \capbest{green} and second-best in \capsecond{yellow}.}
\label{tab:pca_rank_ablation_main}
\begingroup
\setlength{\tabcolsep}{2pt}
\begin{tabular}{@{}l c c c c c c c@{}}
\toprule
\textbf{Method} & \textbf{\# Comp.} & \textbf{RelMSE}$\downarrow$ & \textbf{HR4K} & \textbf{MMStar} & \textbf{MVista} & \textbf{Avg-3} & \textbf{$\Delta$} \\
\midrule
Qwen2.5-VL & -- & -- & 68.13 & 60.46 & 68.50 & 65.70 & 0.00 \\
LH+DA & 3584 & -- & \secondbest{71.88} & 61.80 & 68.90 & 67.53 & +1.83 \\
PCA-0.85+DA & 265 & 0.1494 & 71.00 & \secondbest{62.60} & 69.30 & 67.63 & +1.93 \\
PCA-0.90+DA & 381 & \secondbest{0.0999} & 71.38 & 61.60 & \secondbest{70.10} & \secondbest{67.69} & \secondbest{+1.99} \\
GAP (PCA-0.95+DA) & 629 & \best{0.0500} & \best{73.25} & \best{63.40} & \best{71.00} & \best{69.22} & \best{+3.52} \\
\bottomrule
\end{tabular}
\endgroup
\end{table}

\begin{takeawaybox}
\textbf{Takeaway 3.} PCA-aligned reconstruction outperforms an unconstrained latent head at every tested rank.
\end{takeawaybox}

PCA-aligned reconstruction thus acts as an effective capacity-and-subspace constraint; comparisons with other low-rank parameterizations are left to future work.

\subsection{Generated Latents Contain Task-Relevant Visual Signal}
\label{sec:latent_intervention}

To test whether gains come from content-bearing latents rather than formatting or training-time regularization, we evaluate inference-time interventions on HRBench4K and MathVista.
The zero-latent checkpoint refers to the GAP checkpoint evaluated with latent-token generation suppressed, so no generated PCA-decoded latent embedding is re-injected into the input stream; this row measures training and data effects without inference-time latent feedback.
The noise setting replaces predicted PCA coefficients with Gaussian noise before reconstruction. The clean and noise settings use the same reconstruction and norm-matching rules.

\begin{table}[!ht]
\centering
\caption{Latent content intervention. The zero-latent checkpoint row uses GAP with latent-token generation suppressed, while the noise row replaces predicted PCA coefficients with Gaussian noise before reconstruction. Avg-2 averages HRBench4K Overall and MathVista; $\Delta$ is relative to Qwen2.5-VL-7B Avg-2. Best results are in \capbest{green} and second-best in \capsecond{yellow}.}
\label{tab:latent_intervention}
\begingroup
\setlength{\tabcolsep}{2pt}
\begin{tabular}{@{}l c c l c c c c@{}}
\toprule
\textbf{Intervention} & \textbf{Ckpt.} & \textbf{Latents} & \textbf{Content} & \textbf{HR4K}$\uparrow$ & \textbf{MVista}$\uparrow$ & \textbf{Avg-2}$\uparrow$ & \textbf{$\Delta$} \\
\midrule
Qwen2.5-VL & Base & 0 & None & 68.13 & 68.50 & 68.32 & 0.00 \\
Dense Cap & Dense & 0 & None & 69.38 & 67.70 & 68.54 & +0.22 \\
Zero-latent ckpt. & GAP & 0 & None & \secondbest{71.25} & \secondbest{69.40} & \secondbest{70.33} & \secondbest{+2.01} \\
Noise latents & GAP & 36 & Gauss. noise & 70.38 & 69.00 & 69.69 & +1.37 \\
Clean latents & GAP & 36 & Pred. visual & \best{73.25} & \best{71.00} & \best{72.13} & \best{+3.81} \\
\bottomrule
\end{tabular}
\endgroup
\end{table}

\begin{takeawaybox}
\textbf{Takeaway 4.} Clean generated latents provide useful inference-time visual signal.
\end{takeawaybox}

The zero-latent checkpoint still beats Dense Caption SFT by +1.79 Avg-2, so latent training improves the model even when latent feedback is disabled at inference.
Using clean generated latents further improves Avg-2 from 70.33 to 72.13, while Gaussian coefficient noise drops below both the zero-latent checkpoint and clean-latent setting.
Thus, clean predicted latents carry signal beyond formatting and training-time effects.

The main setting uses 36 latent tokens; Appendix~\ref{sec:token_budget_cost} reports the full token-budget sweep.
Visual latent capacity is therefore most effective when it is aligned, content-bearing, and budgeted rather than simply increased.

\section{Conclusion}
\label{sec:conclusion}

Reusing decoder outputs as visual latents feeds a pre-norm MLLM with inputs it was not trained to consume,
because even after the final RMSNorm the fed-back state is about $4.7\times$ the norm of the input vision embeddings.
GAP removes this mismatch by aligning visual latents at three levels, namely context-grounded latent supervision at the data level,
PCA-aligned reconstruction at the feature level, and difficulty-aware latent assignment at the model level.
On Qwen2.5-VL-7B, GAP is the only latent method in our comparison that improves both perception and reasoning over the base model
(+3.81 Avg-P and +1.35 Avg-R), whereas Monet and LVR gain perception but lose 4.6 and 5.0 points of Avg-R, respectively.
The levels are complementary, since neither feature alignment nor difficulty-aware assignment alone surpasses the base model on reasoning,
while their combination does, and inference-time interventions show that the generated latents carry task-relevant visual signal.
These results suggest that visual latents should be treated as model inputs that must match the input-embedding distribution,
not as raw outputs to recycle.
Our evaluation covers one backbone and five benchmarks with an encoder-specific PCA basis and a shared 95\% retained-variance default, and Appendix~\ref{app:limitations} discusses the remaining limitations, including the latent-token budget trade-off.

\section*{Acknowledgments}
This work was conducted during Yanting Miao's internship at Alibaba Group. We thank Alibaba Group for its support of this research. We also thank Li Xu and Yudong Luo for their feedback on the writing and for early discussions of the idea.
Poupart also acknowledges funding from the Canada CIFAR AI Chair
program, the Natural Sciences and Engineering Research Council of
Canada and a grant from IITP \& MSIT of Korea (No. RS2024-00457882, AI Research Hub Project).

\bibliographystyle{conference}
\bibliography{conference}

\appendix
\clearpage

\section{Architecture Illustration}
\label{app:architecture}

Figure~\ref{fig:architecture} makes the interleaving explicit: ordinary text tokens are decoded through the LM head, latent spans are routed through the PCA-aligned latent head, and each reconstructed visual embedding is fed back as the next latent-token input.

\section{Limitations and Scope}
\label{app:limitations}

\begin{enumerate}[leftmargin=*]
    \item The full training and evaluation pipeline is validated on Qwen2.5-VL 7B and Monet-7B. Extending the same analysis to other pre-norm MLLMs is necessary before claiming cross-backbone universality. Post-norm backbones may still suffer from latent subspace mismatch, but the norm-growth mechanism studied here is specific to pre-norm residual streams. Our benchmark suite follows the perception benchmarks of Monet and LVR and adds MathVista and WeMath for reasoning; broader task coverage remains to be tested.
    \item The LVR-49K control matches the training data and backbone to GAP, while Monet remains a prior-system baseline. The matched-data comparison helps separate the contribution of the curated corpus from that of the overall training method, but it does not isolate PCA or difficulty-aware assignment individually; those effects require the corresponding component ablations.
    \item The difficulty threshold $\tau=0$ is heuristic. The matched LH+PCA versus LH+PCA+DA control holds the training configuration fixed and varies the supervision assignment. Its five evaluator-seed repeats measure evaluation variability on fixed prediction sets, not variability across independently trained models, and therefore do not by themselves establish training-seed robustness of the DA effect. Sensitivity to other difficulty thresholds remains to be studied.
    \item PCA constrains latents to an empirical affine subspace and reduces head capacity. Appendix~\ref{app:pca_rank} adds a component-count sweep, but random-basis, no-mean, and matched-parameter full-head controls remain incomplete. The numerical basis is encoder-specific and must be re-estimated for a new vision encoder or fitting distribution, and the rank selected at a given retained-variance ratio depends on the fitting distribution (Appendix~\ref{app:pca_fit_distribution}). All experiments use 95\% as a shared default. Variance retention also does not guarantee that low-variance but task-relevant directions are preserved.
    \item Gains are not uniform across benchmarks. With the shared 36-token budget, GAP trails the original LVR on MME-RealWorld-Lite, whose remote-sensing, monitoring, and in-the-wild OCR scenarios are underrepresented in our chart-, geometry-, and visual-search-oriented training mixture; LVR-49K shows a similar decrease. Increasing the budget to 144 tokens recovers most of this gap (Appendix~\ref{app:mme_budget}), but no single fixed budget is best on every benchmark, so the token count should be treated as a perception--reasoning and accuracy--cost trade-off.
    \item The \texttt{<parser>} field describes the intended auxiliary visual evidence; it is not a decoding of the continuous latent representation or a guarantee that the latent vector faithfully encodes that description. Stronger parser--latent faithfulness tests are left for future work.
    \item The auxiliary-image requirement limits the approach to domains where meaningful auxiliary views can be generated or sourced and requires careful duplicate and leakage checks against evaluation benchmarks. We have not systematically repeated the full data-construction pipeline with alternative teachers or image generators, so sensitivity to these choices remains unquantified.
    \item Future work should test the same alignment design on additional MLLM backbones and study whether RL-based latent training gives further gains once the feedback space is input-compatible. Our implementation uses a fixed latent-token budget, as in Monet and LVR; allocating latent tokens dynamically per example, e.g., based on task difficulty, image complexity, or generation confidence, is another open direction.
\end{enumerate}

\section{Broader Impact}
\label{app:broader_impact}

GAP aims to make visual latent reasoning more reliable and inspectable by aligning generated latent tokens with input-compatible visual features and by pairing latent supervision with a human-readable parser trace.
This may benefit multimodal systems that need compact intermediate visual reasoning without calling external visual tools or image generators at inference time, potentially reducing pipeline complexity and inference-time resource use.

At the same time, stronger visual reasoning systems can also make incorrect answers appear more grounded.
The \texttt{<parser>} field describes the intended visual evidence, but it is not a formal guarantee that the continuous latent vector faithfully encodes exactly that content.
We therefore view parser traces, intervention tests, and qualitative examples as diagnostic aids rather than complete explanations.
Future deployments should include task-specific validation, leakage checks, and caution in high-stakes settings where visual errors could affect safety, medical, legal, or financial decisions.

\section{External Assets, Licenses, and Terms}
\label{app:asset_licenses}

Table~\ref{tab:asset_licenses} summarizes the existing models, datasets, benchmarks, services, and comparison assets used in this work.
License and terms information is taken from the upstream pages listed in the table.
When an upstream benchmark does not declare a standard license, or declares restrictive research-only terms, we use it only for evaluation or comparison and do not redistribute the benchmark content.
The released latent-supervision data (36,985 examples; Appendix~\ref{app:data}) records the upstream source of every example in a per-example \texttt{data\_source} field, and its dataset card maps each source to the upstream page and license listed in Table~\ref{tab:asset_licenses}.
The release is intended for non-commercial research, and each example remains subject to the license and terms of its upstream source.

\begingroup
\scriptsize
\setlength{\tabcolsep}{3pt}
\renewcommand{\arraystretch}{1.12}
\begin{longtable}{@{}>{\raggedright\arraybackslash}p{0.17\linewidth}
                  >{\raggedright\arraybackslash}p{0.20\linewidth}
                  >{\raggedright\arraybackslash}p{0.27\linewidth}
                  >{\raggedright\arraybackslash}p{0.28\linewidth}@{}}
\caption{External assets, upstream pages, and license or terms-of-use status.}
\label{tab:asset_licenses}\\
\toprule
\textbf{Asset} & \textbf{Role in this work} & \textbf{Upstream page or version} & \textbf{License / terms status and use} \\
\midrule
\endfirsthead
\toprule
\textbf{Asset} & \textbf{Role in this work} & \textbf{Upstream page or version} & \textbf{License / terms status and use} \\
\midrule
\endhead
\midrule
\multicolumn{4}{r}{Continued on next page} \\
\endfoot
\bottomrule
\endlastfoot
Qwen2.5-VL-7B-Instruct & Student backbone and zero-shot baseline & \url{https://huggingface.co/Qwen/Qwen2.5-VL-7B-Instruct} & Apache-2.0. We use the model as the base MLLM and preserve upstream attribution in released model metadata. \\
Qwen3-VL-235B-A22B-Thinking & Offline teacher for response synthesis & \url{https://huggingface.co/Qwen/Qwen3-VL-235B-A22B-Thinking} & Apache-2.0. Used only to synthesize structured supervision; no teacher weights are redistributed. \\
Gemini image-generation service (Nano Banana 2) & Auxiliary-image generator during data construction & \url{https://ai.google.dev/gemini-api/terms} & Governed by Gemini API / Google AI Studio terms and the associated prohibited-use policy. We use generated auxiliary images for research supervision and do not redistribute the service or model weights. \\
Monet-7B and Monet-SFT-125K & Prior latent baseline, data-source comparison, and source of our Visual-CoT, Zebra-CoT, CogCoM, and ReFocus training examples & \url{https://huggingface.co/NOVAglow646/Monet-7B}; \url{https://huggingface.co/datasets/NOVAglow646/Monet-SFT-125K} & Hugging Face model and dataset pages list MIT. We cite the Monet paper and use Monet-7B for baseline comparison. We use Monet-SFT-125K for the data ablation in Table~\ref{tab:data_effectiveness} and as the source of our Visual-CoT, Zebra-CoT, CogCoM, and ReFocus examples, which remain subject to the original dataset terms listed below. \\
LVR & Prior latent-method comparison & \url{https://arxiv.org/abs/2511.19094} & Paper comparison only; we do not redistribute LVR code or model weights. If released upstream assets are used in future versions, their license will be added to this table. \\
HR-Bench / HRBench4K & Perception evaluation benchmark & \url{https://huggingface.co/datasets/DreamMr/HR-Bench} & Upstream Hugging Face metadata lists license as \texttt{other}, and the repository license file is empty. Used only for evaluation; benchmark content is not redistributed. \\
MMStar & Perception evaluation benchmark & \url{https://github.com/MMStar-Benchmark/MMStar}; \url{https://huggingface.co/datasets/Lin-Chen/MMStar} & No standard dataset license is declared on the primary GitHub / Hugging Face pages. Used only for evaluation; benchmark content is not redistributed. \\
MME-RealWorld-Lite & Perception and reasoning evaluation benchmark & \url{https://github.com/MME-Benchmarks/MME-RealWorld}; \url{https://huggingface.co/datasets/yifanzhang114/MME-RealWorld-Lite} & Upstream terms restrict use to academic research and prohibit commercial use and redistribution without approval. We use the Lite split only for evaluation and do not redistribute it. \\
MathVista & Reasoning evaluation benchmark and qualitative example source & \url{https://huggingface.co/datasets/AI4Math/MathVista} & CC-BY-SA-4.0 for new dataset contributions, with source-image copyrights retained by original owners and dataset-specific test-set / no-training conditions. We use it only for evaluation and qualitative analysis. \\
We-Math & Reasoning evaluation benchmark & \url{https://huggingface.co/datasets/We-Math/We-Math} & CC-BY-NC-4.0. Used only for evaluation; benchmark content is not redistributed. \\
POPE & OOD hallucination-sensitive sanity check & \url{https://github.com/RUCAIBox/POPE} & MIT license on the official repository. Used only for evaluation in Appendix~\ref{app:pope_ood}. \\
Visual-CoT & Source pool for visual-CoT-style training examples & \url{https://huggingface.co/datasets/deepcs233/Visual-CoT} & Hugging Face metadata lists Apache-2.0, while the dataset card also states Attribution-NonCommercial 4.0 and research-purpose restrictions. Released examples keep these terms and are treated as CC BY-NC 4.0. \\
Zebra-CoT & Source pool for counting, visual-search, and geometry examples & \url{https://huggingface.co/datasets/multimodal-reasoning-lab/Zebra-CoT} & CC-BY-NC-4.0. Used for research training data construction; redistribution follows the upstream non-commercial terms. \\
CogCoM / CoMDataset & Focused visual reasoning source pool & \url{https://huggingface.co/datasets/qijimrc/CoMDataset}; \url{https://github.com/THUDM/CogCoM} & Dataset contributions are CC-BY-SA-4.0 with original source copyrights retained; CogCoM code is Apache-2.0 and model weights have a separate model license. We use the dataset portion only where terms permit. \\
ReFocus & Structured visual reasoning source pool & \url{https://huggingface.co/datasets/ReFocus/ReFocus_Data} & Apache-2.0. Used for research training data construction with upstream attribution. \\
MathVision & Multimodal math source pool & \url{https://huggingface.co/datasets/MathLLMs/MathVision} & MIT. Used for multimodal math examples in the training mixture with upstream attribution. \\
GeoQA and Geometry3K / Geo3K & Geometry source pools & \url{https://huggingface.co/datasets/hz2475/geoQA}; \url{https://huggingface.co/datasets/xyliu6/geometry3k} & The referenced GeoQA port lists MIT; the referenced Geometry3K port lists Apache-2.0. We include source identifiers in the release manifest and avoid exact evaluation duplicates as described in Appendix~\ref{app:data}. \\
ChartQA & Chart-understanding source pool & \url{https://huggingface.co/datasets/ahmed-masry/ChartQA}; \url{https://github.com/vis-nlp/ChartQA} & GPL-3.0 on the Hugging Face and GitHub pages; chart images were crawled by the ChartQA authors from online sources such as Statista and Pew Research Center, and their copyrights are retained by the original owners. Used for research training data construction with upstream attribution. \\
MM-MATH & Multimodal math source pool & \url{https://huggingface.co/datasets/THU-KEG/MM_Math} & MIT. Problems are collected from Chinese middle-school exams. Used for multimodal math examples in the training mixture with upstream attribution. \\
\end{longtable}
\endgroup

\section{Background: PCA and Pre-Norm Decoders}
\label{app:background}

\textbf{PCA reconstruction.}
Let $\{\mathbf{v}_i\}_{i=1}^{N}$ denote vision embeddings extracted from auxiliary images by the frozen vision encoder, where $\mathbf{v}_i \in \mathbb{R}^{d}$.
We compute the empirical mean and covariance
\begin{equation}
    \boldsymbol{\mu} = \frac{1}{N}\sum_{i=1}^{N}\mathbf{v}_i,\qquad
    \boldsymbol{\Sigma} =
    \frac{1}{N}\sum_{i=1}^{N}(\mathbf{v}_i-\boldsymbol{\mu})(\mathbf{v}_i-\boldsymbol{\mu})^\top .
\end{equation}
Let $\mathbf{P}_k=[\mathbf{p}_1,\ldots,\mathbf{p}_k]\in\mathbb{R}^{d\times k}$ be the top-$k$ eigenvectors of $\boldsymbol{\Sigma}$.
For any centered vision embedding, PCA gives coefficients and a rank-$k$ reconstruction:
\begin{equation}
    \mathbf{c}=\mathbf{P}_k^\top(\mathbf{v}-\boldsymbol{\mu}),\qquad
    \hat{\mathbf{v}}=\mathbf{P}_k\mathbf{c}+\boldsymbol{\mu}.
    \label{eq:app_pca_reconstruct}
\end{equation}
We report the relative reconstruction error as the fraction of centered vision-embedding variance not captured by the rank-$k$ subspace:
\begin{equation}
    \mathrm{RelMSE}(k)
    =
    \frac{
    \frac{1}{N}\sum_{i=1}^{N}
    \left\|
    \mathbf{v}_i -
    \left(\mathbf{P}_k\mathbf{P}_k^\top(\mathbf{v}_i-\boldsymbol{\mu})+\boldsymbol{\mu}\right)
    \right\|_2^2
    }{
    \frac{1}{N}\sum_{i=1}^{N}
    \left\|\mathbf{v}_i-\boldsymbol{\mu}\right\|_2^2
    }
    =
    1-\frac{\sum_{j=1}^{k}\lambda_j}{\sum_{j=1}^{d}\lambda_j},
    \label{eq:relative_reconstruction_error}
\end{equation}
where $\lambda_j$ are the eigenvalues of the empirical covariance.
Lower relative reconstruction error means that the PCA subspace retains more of the auxiliary-image vision-embedding variance.
In our model, the latent head predicts $\mathbf{c}_t$ rather than an unconstrained full-dimensional embedding.
The generated latent is then reconstructed by Eq.~\ref{eq:app_pca_reconstruct}, which constrains feedback tokens to the affine subspace estimated from real vision embeddings.

\begin{figure}[t]
\centering
\includegraphics[width=\linewidth]{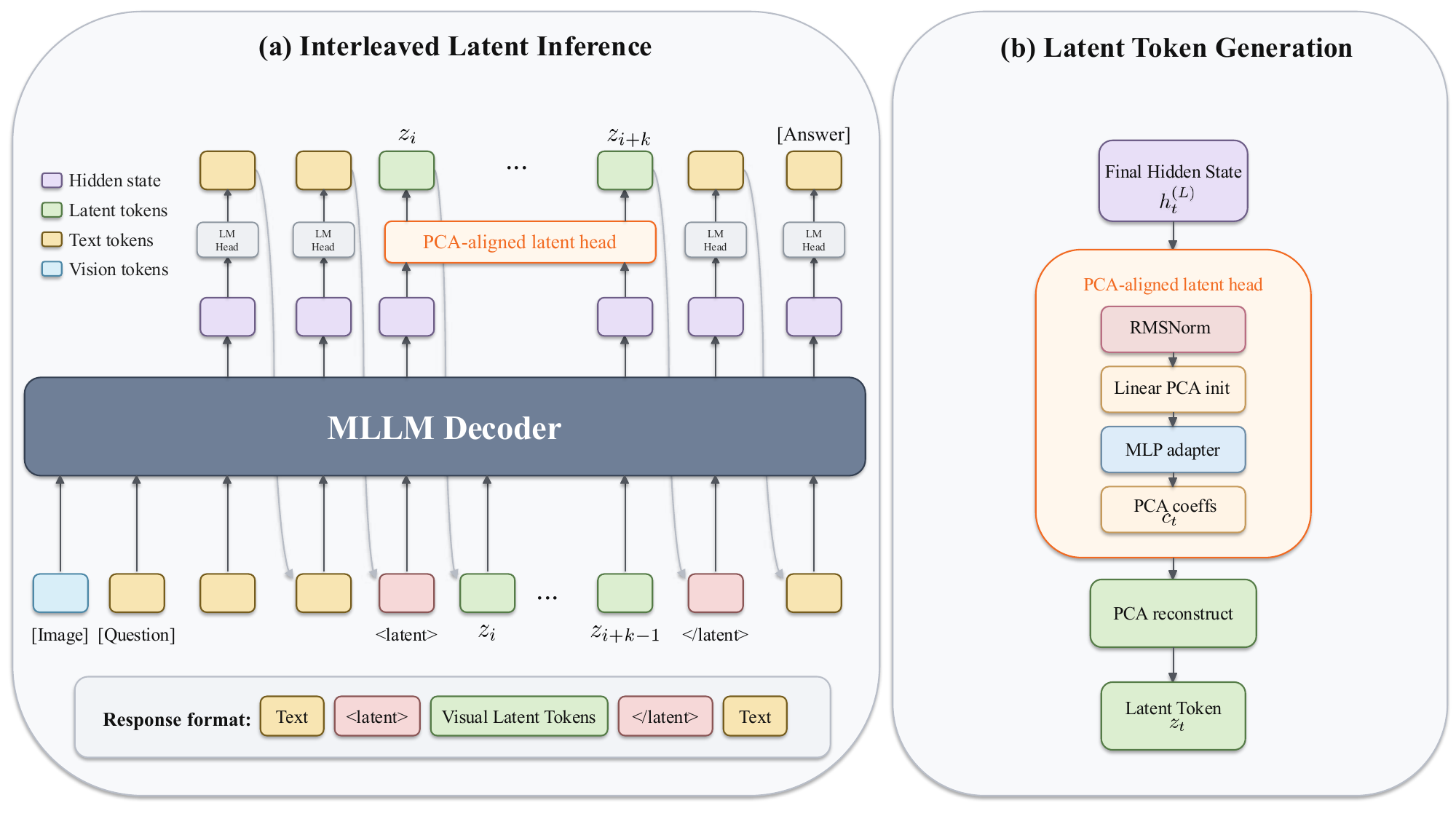}
\caption{Interleaved visual latent inference and latent-token generation.
(a) The MLLM generates ordinary text tokens through the LM head and generates visual latent tokens inside a \texttt{<latent>} span. Each generated latent token is fed back autoregressively as a subsequent latent-token input, giving a response format of text, visual latent tokens, and text.
(b) A latent token is generated from the final decoder state by applying the backbone's native final RMSNorm, a PCA-aligned latent head that predicts PCA coefficients, and a fixed PCA reconstruction step. The reconstructed embedding is then used as the next latent token; auxiliary-image targets are used only to train this path.}
\label{fig:architecture}
\end{figure}

\textbf{Pre-norm decoder states.}
Modern decoder-only transformers, including the Qwen2.5-VL backbone used in our experiments, are typically pre-norm architectures.
Ignoring layer indices on attention keys and values for readability, a pre-norm block has the form
\begin{align}
    \mathbf{x}_{\ell}' &= \mathbf{x}_{\ell}
    + \mathrm{Attn}_{\ell}\!\left(\mathrm{RMSNorm}_{\ell}(\mathbf{x}_{\ell})\right),\\
    \mathbf{x}_{\ell+1} &= \mathbf{x}_{\ell}'
    + \mathrm{MLP}_{\ell}\!\left(\mathrm{RMSNorm}_{\ell}'(\mathbf{x}_{\ell}')\right).
\end{align}
RMSNorm itself rescales a hidden vector by its root-mean-square magnitude:
\begin{equation}
    \mathrm{RMSNorm}(\mathbf{x})
    =
    \mathbf{g}\odot
    \frac{\mathbf{x}}
    {\sqrt{\frac{1}{d}\sum_{j=1}^{d}x_j^2+\epsilon}},
    \label{eq:app_rmsnorm}
\end{equation}
where $\mathbf{g}$ is a learned scale vector.
This placement explains why hidden-state norms can accumulate with depth.
For a single residual update, write
\begin{equation}
    \mathbf{x}_{\ell+1}
    =
    \mathbf{x}_{\ell}
    +
    \mathbf{u}_{\ell},
    \qquad
    \mathbf{u}_{\ell}
    =
    F_{\ell}\!\left(\mathrm{RMSNorm}(\mathbf{x}_{\ell})\right),
\end{equation}
where $F_{\ell}$ denotes the attention or MLP branch.
Then
\begin{equation}
    \left\|\mathbf{x}_{\ell+1}\right\|_2^2
    =
    \left\|\mathbf{x}_{\ell}\right\|_2^2
    +
    \left\|\mathbf{u}_{\ell}\right\|_2^2
    +
    2\left\langle \mathbf{x}_{\ell}, \mathbf{u}_{\ell}\right\rangle .
    \label{eq:app_prenorm_norm_growth}
\end{equation}
Pre-norm controls the input scale seen by $F_{\ell}$, but it does not normalize the residual stream after $\mathbf{u}_{\ell}$ is added.
Therefore, if residual updates have nonzero energy and are not consistently anti-aligned with the current residual stream, i.e.,
\begin{equation}
    \mathbb{E}\!\left[\left\langle \mathbf{x}_{\ell}, \mathbf{u}_{\ell}\right\rangle\right]\approx 0,
    \qquad
    \mathbb{E}\!\left[\left\|\mathbf{u}_{\ell}\right\|_2^2\right] > 0,
\end{equation}
then the expected squared norm accumulates:
\begin{equation}
    \mathbb{E}\!\left[\left\|\mathbf{x}_{L}\right\|_2^2\right]
    \approx
    \mathbb{E}\!\left[\left\|\mathbf{x}_{0}\right\|_2^2\right]
    +
    \sum_{\ell=0}^{L-1}
    \mathbb{E}\!\left[\left\|\mathbf{u}_{\ell}\right\|_2^2\right].
    \label{eq:app_prenorm_accumulation}
\end{equation}
This is not a monotonicity guarantee for every individual layer, because the inner-product term in Eq.~\ref{eq:app_prenorm_norm_growth} can be negative.
Rather, it describes the expected residual-stream accumulation pattern observed in pre-norm decoders: normalization stabilizes each branch input, while the unnormalized residual stream can grow with depth.
Many pre-norm decoders also apply a final RMSNorm before the output readout:
\begin{equation}
    \bar{\mathbf{h}}_t
    =
    \mathrm{RMSNorm}_{\mathrm{final}}(\mathbf{h}_t^{(L)}),\qquad
    \mathrm{logits}_t = \mathbf{W}_{\mathrm{vocab}}\bar{\mathbf{h}}_t .
\end{equation}
Our latent head uses this same backbone-normalized state:
\begin{equation}
    \mathbf{c}_t = \mathcal{F}_{\theta}(\bar{\mathbf{h}}_t),\qquad
    \hat{\mathbf{v}}_t = \mathbf{P}_k\mathbf{c}_t+\boldsymbol{\mu}.
\end{equation}
This distinction is important.
The final RMSNorm stabilizes the decoder readout scale, but it does not make $\bar{\mathbf{h}}_t$ an input-compatible vision embedding.
PCA reconstruction supplies the missing input-compatibility step: it maps the normalized decoder state into coefficients of the empirical vision-embedding subspace before the latent is fed back as the next input token.

\section{PCA Component-Count and Capacity-Control Ablation}
\label{app:pca_rank}

Table~\ref{tab:pca_rank_ablation} varies the PCA retained-variance threshold while keeping the backbone, curated training data, 36 latent tokens, difficulty-aware supervision, and evaluation benchmarks fixed.
The baseline Qwen2.5-VL-7B row has no latent head, while LH+DA uses the full latent head in the original $d=3584$ vision-embedding space without PCA compression.
All PCA rows use the same difficulty-aware policy and differ only in the number of retained PCA components induced by the retained-variance threshold.

\begin{table}[h]
\centering
\caption{PCA component-count ablation under difficulty-aware latent supervision. Component ratio is measured relative to the original vision-embedding dimension $d=3584$; the full-head LH+DA baseline predicts in this original space without PCA compression. RelMSE is the relative reconstruction error in Eq.~\ref{eq:relative_reconstruction_error}. Avg-3 is the unweighted average of HRBench4K Overall, MMStar, and MathVista. $\Delta$ is measured relative to Qwen2.5-VL-7B Avg-3. Best results are in \capbest{green} and second-best in \capsecond{yellow}.}
\label{tab:pca_rank_ablation}
\begingroup
\footnotesize
\setlength{\tabcolsep}{1pt}
\begin{tabular}{@{}l c c c c c c c c c c@{}}
\toprule
\textbf{Run} & \textbf{Var.} & \textbf{Ratio} & \textbf{\# Comp.} & \textbf{RelMSE}$\downarrow$ & \textbf{Tok.} & \textbf{HR4K} & \textbf{MMStar} & \textbf{MVista} & \textbf{Avg-3} & \textbf{$\Delta$} \\
\midrule
Qwen2.5-VL & -- & -- & -- & -- & 0 & 68.13 & 60.46 & 68.50 & 65.70 & 0.00 \\
LH+DA & 1.00 & 100.0\% & 3584 & -- & 36 & \secondbest{71.88} & 61.80 & 68.90 & 67.53 & +1.83 \\
PCA-0.85+DA & 0.85 & 7.4\% & 265 & 0.1494 & 36 & 71.00 & \secondbest{62.60} & 69.30 & 67.63 & +1.93 \\
PCA-0.90+DA & 0.90 & 10.6\% & 381 & \secondbest{0.0999} & 36 & 71.38 & 61.60 & \secondbest{70.10} & \secondbest{67.69} & \secondbest{+1.99} \\
\gapmethod & 0.95 & 17.6\% & 629 & \best{0.0500} & 36 & \best{73.25} & \best{63.40} & \best{71.00} & \best{69.22} & \best{+3.52} \\
\bottomrule
\end{tabular}
\endgroup
\end{table}

The uncompressed LH+DA head already improves over the base model, showing that the auxiliary latent objective is useful.
However, all PCA+DA variants further improve Avg-3 relative to the uncompressed head, indicating that constraining the generated latent to an empirical vision-embedding coordinate system is helpful beyond adding a full latent head.
The relative reconstruction error decreases from 0.1494 at PCA-0.85+DA to 0.0999 at PCA-0.90+DA and 0.0500 at PCA-0.95+DA, matching the expected residual variance left outside the retained PCA subspace.
The downstream trend follows the same broad direction: Avg-3 rises from 67.63 and 67.69 with fewer components to 69.22 at PCA-0.95+DA.
This does not imply that reconstruction error alone determines benchmark performance, but it supports the interpretation that retaining too few components discards task-relevant visual detail.
At the same time, the PCA rows outperform the uncompressed LH+DA head, suggesting that the benefit is not merely from increasing latent-head capacity; the input-space PCA parameterization itself is useful.
We therefore interpret PCA as a subspace-and-capacity design choice rather than as a standalone normalization operation.

\section{OOD Hallucination Evaluation}
\label{app:pope_ood}

We additionally evaluate POPE as an OOD hallucination-sensitive sanity check.
POPE uses yes/no object-recognition queries under popular, adversarial, and random negative sampling, which differs from our main benchmarks: the desired behavior is conservative binary recognition rather than producing richer visual evidence for long-form reasoning.

\begin{table}[h]
\centering
\caption{OOD hallucination evaluation on POPE. Scores are reported for the overall benchmark and its popular, adversarial, and random splits. Best results are in \capbest{green} and second-best in \capsecond{yellow}.}
\label{tab:pope_ood}
\begin{tabular}{l cccc}
\toprule
\textbf{Method} & \textbf{Overall}$\uparrow$ & \textbf{Popular}$\uparrow$ & \textbf{Adversarial}$\uparrow$ & \textbf{Random}$\uparrow$ \\
\midrule
Qwen2.5-VL-7B & \secondbest{86.23} & \best{86.34} & \best{85.33} & 87.03 \\
Ours (LH+DA, no PCA) & 85.72 & 85.70 & 84.36 & \secondbest{87.09} \\
Ours-\gapmethod & \best{86.32} & \secondbest{85.76} & \secondbest{84.49} & \best{88.70} \\
\bottomrule
\end{tabular}
\end{table}

Table~\ref{tab:pope_ood} shows that the unconstrained LH+DA variant slightly underperforms the base model on overall POPE, despite improving the main perception aggregate.
In contrast, GAP recovers this drop and slightly improves the overall score.
We interpret this pattern as suggestive evidence that an unconstrained full latent head can over-specialize to the curated latent-supervision distribution, while the PCA reconstruction path acts as a capacity and subspace constraint by forcing generated latents to lie in the empirical vision-embedding subspace.
This should not be read as a claim that GAP solves hallucination: the overall gain is mainly driven by the random split, while the popular and adversarial POPE splits remain below the base model, indicating that hallucination-sensitive OOD cases remain challenging.

\section{Implementation Details}
\label{app:impl}

\textbf{Training hyperparameters.}
We fine-tune with AdamW, weight decay 0.01, cosine learning-rate decay, warmup ratio 0.03, peak base learning rate $1{\times}10^{-5}$, peak latent-head learning rate $1{\times}10^{-5}$, global batch size 128, per-device batch size 1, and 2 epochs.
The main difficulty-aware run uses 36 latent tokens; Section~\ref{sec:token_budget_cost} reports a latent-token-budget ablation because this choice affects both alignment capacity and inference cost.
Training uses bf16, FlashAttention-2, gradient checkpointing, 16 dataloader workers, and seed 12345.

\textbf{PCA computation.}
After preprocessing the curated 49K corpus, we split the remaining 36,985 examples into training and validation partitions using a 90/10 protocol and fit PCA exclusively on the training partition, which contains 33,279 samples.
No validation examples or downstream benchmark images or embeddings are used to fit PCA.
We encode each auxiliary-image occurrence with the frozen vision encoder and adaptively pool its outputs into fixed token-level embeddings, yielding approximately 992,664 embedding vectors.
We center these vectors, accumulate their covariance matrix, and obtain the principal directions by eigendecomposition.
The minimum rank retaining 95\% of the fitting-data variance is $k=629$.
The empirical mean and PCA basis are stored and frozen during student training; the same fit is reused across checkpoints, random seeds, and downstream tasks sharing the same vision encoder and fitting data.
The basis is encoder-specific.
We do not refit PCA separately for individual examples, checkpoints, or evaluation benchmarks.

\textbf{Latent head implementation.}
The latent projector is initialized from $\mathbf{P}_k^\top$ so that the head starts from the empirical vision subspace rather than an unconstrained random basis.
We then refine this projection with a compact SwiGLU adapter.
At inference time, ordinary text positions use the LM head, whereas latent positions use the latent head to produce PCA coefficients that are decoded by the fixed basis before feedback, as illustrated in Appendix Figure~\ref{fig:architecture}.

\textbf{Scheduled sampling.}
During teacher-forced latent training, \texttt{<|latent\_pad|>} positions receive the auxiliary-image vision features.
When scheduled sampling is enabled as an ablation, these features can be replaced by the model's PCA-decoded predictions before a second forward pass, making training closer to inference where generated latents are fed back autoregressively.
Unless otherwise stated, all results in the main tables use teacher forcing and do not use scheduled sampling.

\textbf{Difficulty-aware data loading.}
For examples with an empirical \texttt{accuracy} field greater than $\tau$, the loader strips the latent span and auxiliary-image placeholder targets, so the example contributes only language-modeling loss.
For examples with \texttt{accuracy} at or below $\tau$, the loader retains the latent span and auxiliary-image target embeddings for joint LM and latent-head supervision.
The main experiments use $\tau=0$, matching Eq.~\ref{eq:difficulty_aware}.

\textbf{EMA norm calibration.}
The EMA norm calibration is used only as a training-free diagnostic for the norm-mismatch analysis.
At inference, the running average $\bar{n}_{\text{EMA}}$ is initialized from the query image's vision embedding norm and updated token by token before rescaling generated latents to the input vision-embedding norm regime.

\section{Training and Inference Cost}
\label{app:latency}

\textbf{Training cost.}
The main \gapmethod\ model is trained on 32 GPUs for 3 hours and 55 minutes of wall-clock time.
This reports student fine-tuning cost only; offline auxiliary-image generation, teacher response generation, PCA computation, and benchmark evaluation are excluded.

\textbf{Offline data-construction cost.}
For the 49K corpus, auxiliary-image generation with Nano Banana 2 is estimated to cost approximately US\$3.31K at the standard API prices used for this estimate.
Teacher responses are synthesized offline with Qwen3-VL-235B-A22B-Thinking.
Measured throughput on eight GPUs gives an estimated cost of approximately 274 GPU-hours for synthesizing 49K responses.
These figures estimate the two data-generation stages rather than a measured end-to-end construction cost.

\textbf{Latency measurement.}
We report wall-clock latency in seconds per sample measured inside the VLMEvalKit inference wrapper.
All rows use the direct benchmark prediction files and ignore duplicate judge/output files.
Because Dense Caption SFT uses the same 49K curated training data and decoding setup but removes latent tokens and auxiliary-image alignment, we use it as the primary latency baseline for relative cost.
Qwen2.5-VL-7B is kept as a performance reference in the main tables, but not as the relative-latency denominator because its earlier timing was collected under an unmatched setup.

\begin{table}[H]
\centering
\caption{Inference cost on HRBench4K and MathVista. Latency is wall-clock seconds per sample. Mean latency is the primary cost metric; relative latency uses Avg-2 mean latency and Dense Caption SFT as the denominator. Avg-2 performance averages HRBench4K Overall and MathVista. For latency and relative cost, lower is better; for Avg-2, higher is better. Best results are in \capbest{green} and second-best in \capsecond{yellow}.}
\label{tab:latency_appendix}
\begingroup
\setlength{\tabcolsep}{4pt}
\begin{tabular}{@{}l c c c c@{}}
\toprule
\textbf{Method} & \textbf{Tok.} & \textbf{Mean HR/MV/Avg-2 (s)} & \textbf{Rel.} & \textbf{Avg-2} \\
\midrule
Dense Caption SFT & 0 & \best{18.81 / 28.74 / 23.78} & \best{1.00$\times$} & 68.54 \\
Ours-\gapmethod & 4 & \secondbest{27.05 / 44.08 / 35.56} & \secondbest{1.50$\times$} & 70.64 \\
Ours-\gapmethod & 16 & 29.55 / 46.50 / 38.02 & 1.60$\times$ & \best{72.27} \\
Ours-\gapmethod & 36 & 29.59 / 47.13 / 38.36 & 1.61$\times$ & \secondbest{72.13} \\
Ours-\gapmethod & 64 & 28.47 / 49.14 / 38.80 & 1.63$\times$ & 71.98 \\
Ours-\gapmethod & 144 & 28.09 / 46.96 / 37.53 & 1.58$\times$ & 70.83 \\
\bottomrule
\end{tabular}
\endgroup
\end{table}

The cost increase is moderate but non-negligible: the main 36-token setting is $1.61\times$ slower than Dense Caption SFT on Avg-2 mean latency, while improving Avg-2 performance from 68.54 to 72.13.
The latency is not strictly monotonic in the latent-token budget because wall-clock time is also affected by generated response length and benchmark-specific long tails.
For example, Dense Caption SFT has Avg-2 P50/P90 latencies of 10.63/30.27 seconds, while the 36-token setting has 16.10/106.71 seconds; the long tail is dominated by MathVista responses.
Thus, latent tokens provide a clear accuracy-cost tradeoff rather than a free improvement, which motivates the budgeted-latent analysis in Figure~\ref{fig:latent_token_budget}.

\section{Latent-Token Budget Sweep}
\label{sec:token_budget_cost}

The number of latent tokens is a core design choice: too few tokens may under-represent auxiliary visual evidence, while too many tokens can introduce unnecessary computation and redundant self-generated visual features.
We evaluate the \gapmethod\ model with different latent-token budgets on HRBench4K, MMStar, and MathVista, using the zero-shot Qwen2.5-VL 7B model as a reference baseline.
Because latent features are arranged as a square grid with equal height and width, all positive token budgets are perfect squares.

\begin{figure}[H]
\centering
\includegraphics[width=0.96\textwidth]{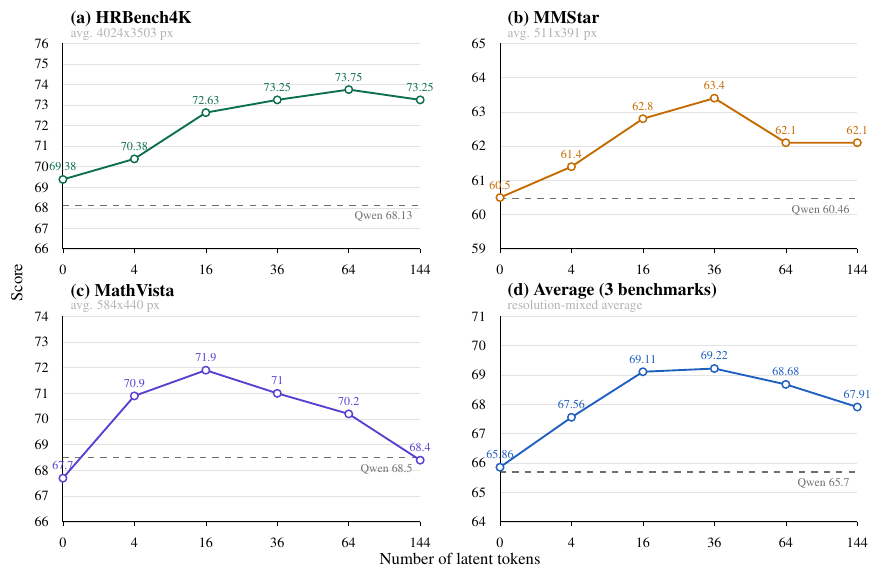}
\caption{Effect of latent-token budget. We sweep the number of generated visual latent tokens while keeping the backbone, training data, PCA-aligned latent head, and difficulty-aware rule fixed. The gray dashed line is the Qwen2.5-VL 7B baseline. Panel subtitles report average image resolution estimated from the evaluated benchmark images. Positive token counts are square latent grids: 4=$2{\times}2$, 16=$4{\times}4$, 36=$6{\times}6$, 64=$8{\times}8$, and 144=$12{\times}12$.}
\label{fig:latent_token_budget}
\end{figure}

Table~\ref{tab:latent_budget_seed_std} reports three-evaluator-seed mean $\pm$ standard deviation results for the token-budget sweep on HRBench4K, MMStar, and MathVista.
These runs cover the 36-token setting used in the main tables and nearby capacity settings. The zero-token row in this sweep is reported separately from the fixed-checkpoint feedback-suppression intervention in Table~\ref{tab:latent_intervention}; the two rows should not be treated as interchangeable controls. Appendix Table~\ref{tab:mme_budget} extends the capacity analysis to MME-RW-Lite.

\begin{table}[H]
\centering
\caption{Three-evaluator-seed mean $\pm$ standard deviation results for the latent-token budget sweep in Figure~\ref{fig:latent_token_budget}. Per-benchmark columns report mean $\pm$ standard deviation, while Avg-3 reports the mean average over HRBench4K Overall, MMStar, and MathVista. Standard deviations reflect evaluator variability, not training-run variability. Best results are in \capbest{green} and second-best in \capsecond{yellow}.}
\label{tab:latent_budget_seed_std}
\begingroup
\footnotesize
\setlength{\tabcolsep}{3pt}
\begin{tabular}{c c c c c}
\toprule
\textbf{Latent tokens} & \textbf{HRBench4K} & \textbf{MMStar} & \textbf{MathVista} & \textbf{Avg-3} \\
\midrule
0   & $69.38{\pm}0.211$ & $60.50{\pm}0.431$ & $67.70{\pm}0.183$ & 65.86 \\
16  & $72.63{\pm}0.176$ & \secondbest{$62.80{\pm}0.387$} & \bestmath{$71.90{\pm}0.215$} & \secondbest{69.11} \\
36  & \secondbest{$73.25{\pm}0.220$} & \bestmath{$63.40{\pm}0.340$} & \secondbest{$71.00{\pm}0.120$} & \best{69.22} \\
64  & \bestmath{$73.75{\pm}0.295$} & $62.10{\pm}0.289$ & $70.20{\pm}0.141$ & 68.68 \\
144 & \secondbest{$73.25{\pm}0.134$} & $62.10{\pm}0.412$ & $68.40{\pm}0.167$ & 67.91 \\
\bottomrule
\end{tabular}
\endgroup
\end{table}

Figure~\ref{fig:latent_token_budget} shows that latent-token capacity helps, but the effect is not monotonic.
The best average over HRBench4K, MMStar, and MathVista is achieved at 36 tokens, with 16 tokens close behind, after which performance saturates or declines.
HRBench4K benefits from larger latent grids: its score rises from 69.38 with zero latent tokens to 73.75 with 64 tokens, and remains high at 144 tokens.
The lower-resolution benchmarks saturate earlier: MathVista peaks at 16 latent tokens, and MMStar peaks around 36 tokens before dropping at larger budgets.
These results suggest that visual latent tokens should be budgeted rather than maximized.

\section{Data Construction Details}
\label{app:data}

\begin{figure}[H]
\centering
\includegraphics[width=0.9\linewidth]{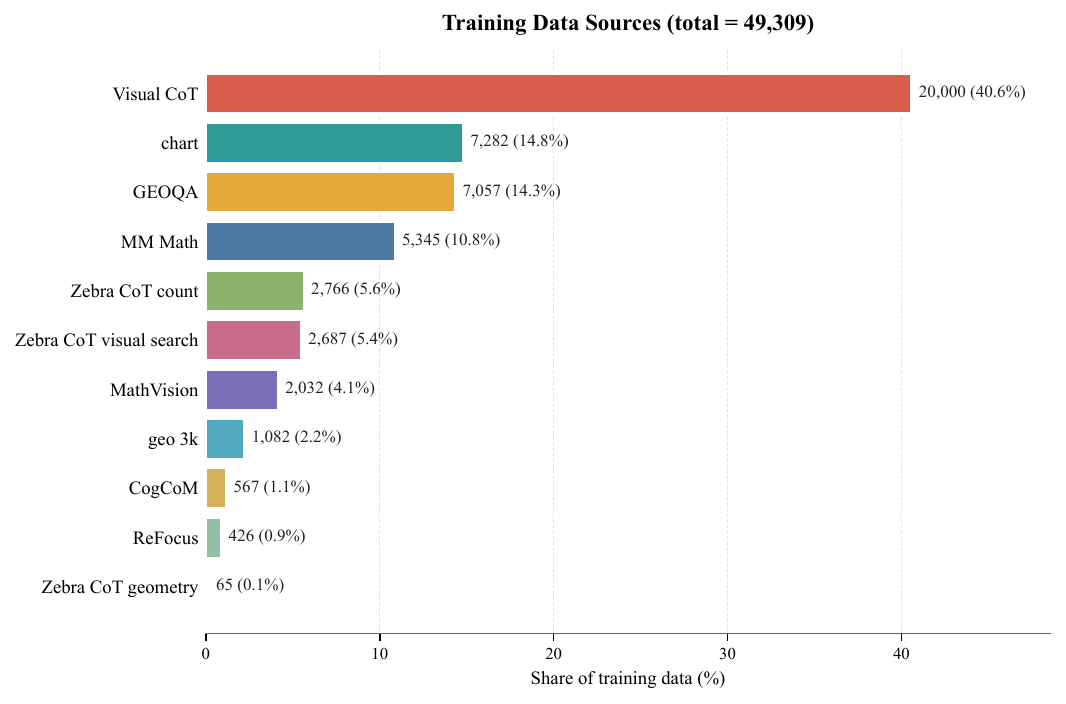}
\caption{Composition of the 49K curated corpus before preprocessing (49,309 examples). Visual CoT is the largest source, followed by chart, GEOQA, and multimodal math data; the remaining sources emphasize geometry, counting, visual search, and focused visual reasoning.}
\label{fig:data_composition}
\end{figure}

\textbf{Auxiliary image generation and response synthesis.}
For each training query, we first use Nano Banana 2 to generate an intermediate auxiliary image intended to expose the missing visual evidence needed by the question.
We then provide the query image, question, and generated auxiliary image to Qwen3-VL-235B-A22B-Thinking, which produces the structured response containing the reasoning trace, \texttt{<latent>} span, \texttt{<parser>} description, and final answer.
Only the student model is trained: the auxiliary image is converted into frozen-ViT features for latent-head supervision, while Qwen3-VL-235B-A22B-Thinking serves only as an offline teacher for response synthesis.

\textbf{Latent serialization.}
The student input contains only the query image and question.
The target response follows the structure \texttt{<think>} context \texttt{<latent>} auxiliary-image embedding targets \texttt{</latent>} \texttt{<parser>} dense caption \texttt{</parser>} continued reasoning \texttt{</think>} \texttt{<answer>} answer \texttt{</answer>}.
In the implementation, each latent span is serialized as \texttt{<|latent\_start|>} followed by fixed \texttt{<|latent\_pad|>} tokens and \texttt{<|latent\_end|>}.
The main setting uses 36 latent tokens unless stated otherwise.

\textbf{Quality filtering and supervision assignment.}
Quality filtering removes examples with degenerate auxiliary-image embeddings or generated \texttt{<parser>} outputs that fail a length or coherence check.
Difficulty-aware assignment is a separate step: examples whose repeated-sampling accuracy exceeds $\tau$ are retained with text-only supervision, rather than removed from the corpus.
For these examples, the loader strips the latent span and auxiliary-image targets, as described in Appendix~\ref{app:impl}.
The curated corpus contains 49,309 examples.
After preprocessing, 36,985 examples remain; we then split them 90/10 into 33,279 training and 3,706 validation examples, which form the released dataset.

\textbf{Leakage audit.}
We audit the curated training set against the evaluation benchmarks used in this paper.
The audit finds no exact image duplicates and no exact normalized question-text duplicates between training and evaluation examples.
We also manually inspect high-similarity cases surfaced by permissive near-duplicate heuristics.
These cases primarily arise from public geometry sources such as GEOQA and Geo3K, which are also reused or adapted by existing evaluation benchmarks.
We therefore avoid claiming that the training and evaluation distributions are fully disjoint, but verify that the training set does not contain exact copies of the evaluated examples.

\subsection{Auxiliary-Image Supervision Example}
\label{app:data_example}

\begin{figure}[H]
\centering
\begin{subfigure}[t]{0.47\linewidth}
    \centering
    \includegraphics[width=\linewidth]{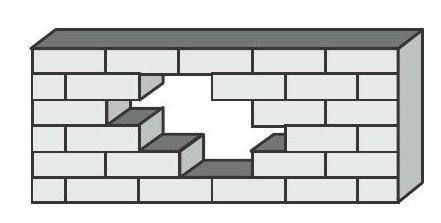}
    \caption{Query image}
\end{subfigure}
\hfill
\begin{subfigure}[t]{0.47\linewidth}
    \centering
    \includegraphics[width=\linewidth]{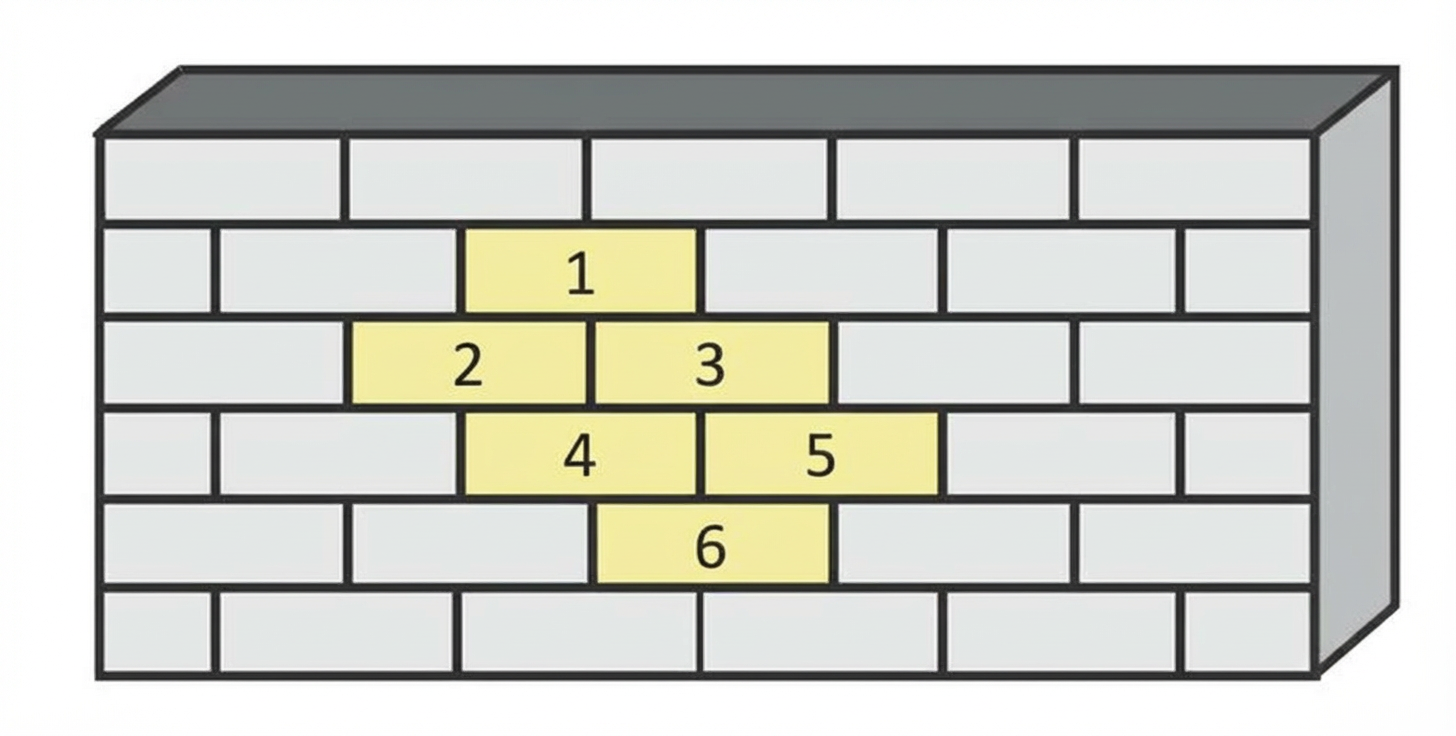}
    \caption{Generated auxiliary image}
\end{subfigure}
\caption{Example of auxiliary-image supervision. The auxiliary image makes the missing visual evidence explicit by reconstructing the wall pattern and numbering the six missing bricks.}
\label{fig:data_example_bricks}
\end{figure}

Figure~\ref{fig:data_example_bricks} shows one training example from our data construction pipeline.
The original query asks: \emph{``How many bricks are missing in the wall?''}
The auxiliary image reconstructs the expected running-bond pattern and labels the six missing brick positions.
This example illustrates that the teacher response provides not only a final answer, but also a human-readable description of the intended auxiliary visual evidence used to supervise the latent channel.

\begin{responsebox}{Qwen3-VL-235B-A22B-Thinking: teacher response excerpt}
To determine how many bricks are missing, I need to reconstruct the complete wall pattern and count the gaps.
The wall uses a standard running bond pattern where each brick is half-offset from the row above and below.
First, I should create an auxiliary image that overlays the missing sections with the expected brick positions to clearly identify each missing brick.

\medskip
\noindent
\texttt{<|vision\_start|><|image\_pad|><|vision\_end|>}

\medskip
\noindent
\texttt{<parser>}
\begin{itemize}[leftmargin=1.2em, itemsep=0.15em, topsep=0.2em]
\item Draw the original brick wall with the damaged section.
\item In the damaged area, add dashed outlines of the bricks that should be present to complete the running bond pattern.
\item Label each missing brick location with a number, 1 through 6, corresponding to the auxiliary image.
\item Highlight the damaged area with a light red background to distinguish it from the intact wall.
\item Ensure the dashed outlines match the size and orientation of standard bricks in the wall.
\end{itemize}
\texttt{</parser>}

\medskip
\noindent
Using the pattern established in the auxiliary image, I can now map the damaged section to see how many bricks should be there.
The damaged area spans multiple rows:
\begin{itemize}[leftmargin=1.2em, itemsep=0.15em, topsep=0.2em]
\item In the row containing brick 1, there is one missing brick.
\item In the row containing bricks 2 and 3, there are two missing bricks.
\item In the row containing bricks 4 and 5, there are two missing bricks.
\item In the row containing brick 6, there is one missing brick.
\end{itemize}
Adding these up gives $1+2+2+1=6$ missing bricks.
The auxiliary image shows this by labeling exactly six missing brick positions.

\medskip
\noindent
\texttt{<answer>}
The wall follows a running bond pattern, where adjacent rows are offset by half a brick.
By comparing the damaged region to the intact rows and the auxiliary image, the missing bricks are labeled 1 through 6.
Therefore, the total number of missing bricks is $\boxed{6}$.
\texttt{</answer>}
\end{responsebox}

This example shows how the teacher turns an image-level counting problem into a concrete visual reconstruction target.
The auxiliary image supplies frozen-ViT features for latent supervision, while the \texttt{<parser>} text records the intended visual content of that latent target.
During student training, the auxiliary image is not supplied as an image input to the student. Its frozen-ViT features fill the teacher-forced \texttt{<|latent\_pad|>} positions and also serve as supervision targets for the latent head.

\section{Qualitative Inference Example}
\label{app:case_study}

\begin{figure}[H]
\centering
\begin{subfigure}[t]{0.30\linewidth}
    \centering
    \includegraphics[width=\linewidth]{figures/mathvista_143_question.jpg}
    \caption{Question image}
\end{subfigure}
\hfill
\begin{subfigure}[t]{0.30\linewidth}
    \centering
    \includegraphics[width=\linewidth]{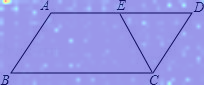}
    \caption{Qwen2.5-VL-7B attention}
\end{subfigure}
\hfill
\begin{subfigure}[t]{0.30\linewidth}
    \centering
    \includegraphics[width=\linewidth]{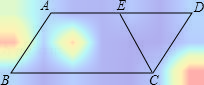}
    \caption{Ours attention}
\end{subfigure}
\caption{A qualitative MathVista example. Attention maps are shown only for qualitative inspection.}
\label{fig:mathvista_case_study}
\end{figure}

Figure~\ref{fig:mathvista_case_study} shows a MathVista geometry example where Qwen2.5-VL-7B predicts an incorrect answer, while our model recovers the correct auxiliary construction and answer.
The question asks:
\emph{``As shown in the figure, in the parallelogram $ABCD$, $CE$ bisects $\angle BCD$ and it intersects the $AD$ edge at point $E$, and $DE=3.0$, then the length of $AB$ is ().''}
The answer choices are: (A)~1, (B)~2, (C)~3, and (D)~6.

\textbf{Baseline failure.}
Qwen2.5-VL-7B retrieves a plausible theorem but grounds it to the wrong geometric configuration.
The following response excerpt is edited only for formatting and math typesetting, while preserving the model's reasoning chain:

\begin{responsebox}{Qwen2.5-VL-7B: incorrect response excerpt}
To solve this problem, we need to use the properties of a parallelogram and the angle bisector theorem.

\begin{enumerate}[leftmargin=1.2em, itemsep=0.2em, topsep=0.3em]
\item \textbf{Identify the properties of the parallelogram.}
In a parallelogram, opposite sides are equal in length, opposite angles are equal, and consecutive angles are supplementary.

\item \textbf{Use the angle bisector theorem.}
The angle bisector theorem states that if a point divides one side of a triangle into segments proportional to the other two sides, then it is the angle bisector of the opposite angle.

\item \textbf{Apply the angle bisector theorem to triangle $BCD$.}
Since $CE$ bisects $\angle BCD$, it divides $\angle BCD$ into two equal angles.
By the angle bisector theorem, the ratio of the segments created by the angle bisector is equal to the ratio of the other two sides.

\item \textbf{Set up the proportion.}
Let $AB=x$ and $CD=x$, since opposite sides of a parallelogram are equal.
According to the angle bisector theorem,
\[
    \frac{DE}{AE} = \frac{CD}{AB}.
\]

\item \textbf{Substitute the known values.}
We have $DE=3$, $AE=AD-DE=x-3$, $CD=x$, and $AB=x$.
Therefore,
\[
    \frac{3}{x-3} = \frac{x}{x}.
\]

\item \textbf{Solve the proportion.}
\[
    \frac{3}{x-3}=1,\qquad 3=x-3,\qquad x=6.
\]
\end{enumerate}

Therefore, the length of $AB$ is $6$.
The correct answer is $\boxed{D}$.
\end{responsebox}

The response applies an angle-bisector-theorem proportion,
$DE/AE=CD/AB$, and then substitutes $AE=AD-DE=x-3$ to predict $AB=6$.
It therefore selects choice D.
This is geometrically invalid: $E$ lies on $AD$, not on the opposite side of $\triangle BCD$, so the stated angle-bisector proportion does not follow from the diagram.
This failure is representative of textually fluent but visually misgrounded reasoning: the model selects a familiar theorem, but attaches it to the wrong line segment.

\textbf{Our response.}
Our model instead identifies the equal-angle relation induced by the parallelogram structure and the angle bisector.
The following response excerpt keeps the main reasoning, latent span, parser content, and final answer:

\begin{responsebox}{Ours: correct response excerpt}
We are given a parallelogram $ABCD$ with $CE$ bisecting $\angle BCD$ and intersecting $AD$ at $E$.
We know $DE=3.0$ and need to find $AB$.
In a parallelogram, opposite sides are equal, so $AB=CD$.
The key is to relate $DE$ to $CD$ using the angle bisector and the parallel lines.

First, recall that in a parallelogram, $AD \parallel BC$.
Therefore, $\angle DEC=\angle ECB$ by alternate interior angles for parallel lines cut by the transversal $CE$.
Since $CE$ bisects $\angle BCD$, $\angle BCE=\angle DCE$.
Combining these angle equalities gives $\angle DEC=\angle DCE$.
Thus, $\triangle DEC$ is isosceles, and $DE=DC$.
The problem states $DE=3.0$, so $DC=3.0$.
Since opposite sides of a parallelogram are equal, $AB=CD=3.0$.

To make this reasoning explicit, we should construct an auxiliary image that highlights the key angle equalities and the isosceles triangle.

\medskip
\noindent
\texttt{<latent>} visual latent tokens \texttt{</latent>}

\medskip
\noindent
\texttt{<parser>}

\begin{itemize}[leftmargin=1.2em, itemsep=0.15em, topsep=0.2em]
\item Draw parallelogram $ABCD$ with $AB \parallel CD$ and $AD \parallel BC$.
\item Mark point $E$ on $AD$ such that $DE=3.0$.
\item Draw segment $CE$.
\item Label $\angle BCE$ and $\angle DCE$ as equal.
\item Label $\angle DEC$ and $\angle DCE$ as equal to show the isosceles triangle.
\item Highlight $\triangle DEC$ with a light fill.
\item Mark $DE=3.0$ on the segment.
\item Indicate with equal-length ticks that $DE=DC$.
\item Write $AB=CD$ near the top side.
\end{itemize}

\texttt{</parser>}

\medskip
\noindent
From the auxiliary image, we see that $\angle DEC=\angle DCE$, so $\triangle DEC$ is isosceles with $DE=DC$.
Given $DE=3.0$, it follows that $DC=3.0$.
Since $ABCD$ is a parallelogram, $AB=CD$.
Therefore, $AB=3.0$.

\medskip
\noindent
\texttt{<answer>}
In parallelogram $ABCD$, $AD \parallel BC$, so $\angle DEC=\angle ECB$.
Since $CE$ bisects $\angle BCD$, $\angle ECB=\angle DCE$.
Therefore, $\angle DEC=\angle DCE$, making $\triangle DEC$ isosceles with $DE=DC$.
Given $DE=3.0$, we have $DC=3.0$.
In a parallelogram, opposite sides are equal, so $AB=CD$.
Thus, the length of $AB$ is $\boxed{3.0}$, corresponding to choice C.
\texttt{</answer>}
\end{responsebox}

\textbf{Interpretation.}
The example illustrates the intended role of visual latent reasoning.
The base model follows a language-plausible but geometrically invalid theorem application, whereas our model produces a latent/parser trace that corresponds to a task-relevant auxiliary construction: highlight $\triangle DEC$, expose $\angle DEC=\angle DCE$, and infer $DE=DC$.
The attention maps provide a qualitative visualization of the two runs, but we do not treat attention as causal evidence.
Instead, this example complements the intervention study in Table~\ref{tab:latent_intervention}, where disabling or replacing the latent rollout reduces performance relative to clean generated latents.

\section{Supplemental Controlled Analyses}
\label{app:rebuttal_controls}

\subsection{Controlled Attribution Summary}
\label{app:attribution}

Table~\ref{tab:attribution} collects the matched comparisons that separate the contribution of the curated data from that of the training method.
Each row changes one factor while keeping the backbone fixed.
The curated data and the training method both contribute; among the method components, PCA alignment gives the largest gains in both aggregates (up to +1.66 Avg-P and +3.92 Avg-R), and DA adds further gains once PCA is in place.

\begin{table}[!ht]
\centering
\small
\caption{Controlled attribution of GAP's gains. Each comparison changes one factor while keeping the Qwen2.5-VL-7B backbone fixed; differences are computed from Tables~\ref{tab:main_perception} and~\ref{tab:main_reasoning}.}
\label{tab:attribution}
\setlength{\tabcolsep}{4pt}
\begin{tabular}{@{}l l c c@{}}
\toprule
\textbf{Factor examined} & \textbf{Controlled comparison} & \textbf{$\Delta$Avg-P} & \textbf{$\Delta$Avg-R} \\
\midrule
Curated 49K data & LVR $\rightarrow$ LVR-49K & +0.10 & +2.02 \\
Method, data fixed & LVR-49K $\rightarrow$ GAP & +0.61 & +4.28 \\
Latent supervision & Dense Cap $\rightarrow$ LH & +0.22 & +1.91 \\
PCA alignment & LH $\rightarrow$ LH+PCA & +1.01 & +3.33 \\
PCA alignment, DA fixed & LH+DA $\rightarrow$ GAP & +1.66 & +3.92 \\
Difficulty-aware assignment & LH+PCA $\rightarrow$ GAP & +1.04 & +1.49 \\
\bottomrule
\end{tabular}
\end{table}

\subsection{Difficulty-Aware Assignment under Evaluator Variability}
\label{app:da_evaluator}

To isolate difficulty-aware assignment (DA) within the PCA-aligned recipe, we compare LH+PCA with LH+PCA+DA while holding the remaining training configuration fixed.
Predictions are regenerated from the same checkpoints using a different GPU configuration from that used for the main tables, and each model's fixed predictions are scored with five evaluator seeds.
Table~\ref{tab:da_evaluator} reports the resulting mean and standard deviation.
GAP achieves a higher mean on all seven metrics in this comparison.
These standard deviations reflect evaluator randomness on fixed outputs, not variability across training runs.

\begin{table}[!ht]
\centering
\small
\caption{Difficulty-aware assignment with fixed prediction sets, evaluated using five evaluator seeds. Values are mean $\pm$ standard deviation in percentage points. These statistics describe evaluator variability, not training-run variability.}
\label{tab:da_evaluator}
\begin{tabular}{@{}lcc@{}}
\toprule
Metric & LH+PCA & GAP (LH+PCA+DA) \\
\midrule
HRBench4K Overall & $72.25 \pm 0.1581$ & $73.53 \pm 0.1559$ \\
HRBench4K FSP     & $87.75 \pm 0.1574$ & $87.80 \pm 0.1418$ \\
HRBench4K FCP     & $56.75 \pm 0.1688$ & $59.25 \pm 0.1620$ \\
MMStar           & $62.21 \pm 0.1730$ & $63.16 \pm 0.1896$ \\
MathVista        & $70.06 \pm 0.2348$ & $71.20 \pm 0.1643$ \\
WeMath-S         & $35.03 \pm 0.1987$ & $36.67 \pm 0.1348$ \\
WeMath-L         & $52.12 \pm 0.2080$ & $56.49 \pm 0.1141$ \\
\bottomrule
\end{tabular}
\end{table}

\subsection{PCA Fitting Distribution and Variance Retention}
\label{app:pca_fit_distribution}

We examine PCA sensitivity using two equally sized, training-only fitting subsets: Global-7334 contains diverse examples, whereas Chart-7334 contains chart examples.
Each subset contains 7,334 samples.
Student training data and the remaining configuration are held fixed within this study; we vary the PCA fitting subset and retained-variance threshold.
Table~\ref{tab:pca_fit_distribution} reports the minimum rank selected by each threshold and the downstream scores.
No validation or downstream benchmark images or embeddings are used to fit these bases.

\begin{table}[!ht]
\centering
\small
\setlength{\tabcolsep}{4pt}
\caption{PCA sensitivity to fitting-data distribution and retained variance. Global-7334 and Chart-7334 each contain 7,334 training samples. Scores are percentages; Avg-2 averages HRBench4K Overall and MathVista. These subset-based ranks differ from the default rank of 629 fitted on the full preprocessed training partition.}
\label{tab:pca_fit_distribution}
\begin{tabular}{@{}llrrrrrr@{}}
\toprule
Fitting set & Variance & Rank & HR Ovr. & FSP & FCP & MathVista & Avg-2 \\
\midrule
Global-7334 & 85\% & 163  & 67.38 & 83.30 & 51.45 & 65.48 & 66.43 \\
Chart-7334  & 85\% & 71   & 68.94 & 83.75 & 54.13 & 65.75 & 67.35 \\
Global-7334 & 90\% & 246  & 69.30 & 84.20 & 54.40 & 66.88 & 68.09 \\
Chart-7334  & 90\% & 114  & 69.50 & 84.25 & 54.75 & 65.60 & 67.55 \\
Global-7334 & 95\% & 420  & 70.92 & 86.25 & 55.58 & 66.53 & 68.73 \\
Chart-7334  & 95\% & 216  & 69.50 & 84.50 & 54.50 & 65.70 & 67.60 \\
Global-7334 & 99\% & 1104 & 69.63 & 84.13 & 55.13 & 66.60 & 68.12 \\
Chart-7334  & 99\% & 618  & 69.75 & 84.25 & 55.25 & 65.00 & 67.38 \\
\bottomrule
\end{tabular}
\end{table}

Both fitting subsets obtain their highest Avg-2 at 95\%, although the selected rank differs substantially (420 for Global-7334 vs.\ 216 for Chart-7334).
The margin over nearby thresholds also depends on the fitting distribution: Global-90 is 0.64 points below Global-95 in Avg-2, whereas Chart-90 is only 0.05 points below Chart-95.
At equal fitting-set size, Global-95 exceeds the best Chart setting on all reported metrics, so the fitting distribution affects downstream accuracy even at a fixed threshold.
Our main model uses one frozen basis and a shared 95\% rule across benchmarks; this study does not entail benchmark-specific PCA fitting.
The numerical basis remains specific to its vision encoder and fitting distribution, and should be re-estimated when those change.
In particular, ranks 420 and 216 at 95\% in this study should not be confused with rank 629 from the full preprocessed training partition.

\subsection{Latent-Token Capacity on MME-RealWorld-Lite}
\label{app:mme_budget}

Table~\ref{tab:mme_budget} examines whether increasing the latent-token budget improves the MME-RealWorld-Lite results.
Increasing the budget from 36 to 144 tokens raises Overall, Reasoning, and Perception by 2.39, 2.13, and 2.56 points, respectively, without MME-specific training data or PCA refitting.
The 144-token model exceeds LVR on Overall and Reasoning and remains 0.20 points below LVR on Perception.
These results show a capacity trade-off within the evaluated setting, rather than a uniform ranking across all budgets and metrics.
We retain 36 tokens as the shared default used to balance perception and reasoning across benchmarks; the 144-token result is a capacity analysis, not a replacement selected separately for each benchmark.
Both settings use a fixed token budget and do not implement dynamic per-example allocation.

\begin{table}[!ht]
\centering
\small
\caption{Latent-token capacity on MME-RealWorld-Lite. Scores are percentages. The default GAP configuration uses 36 tokens; the 144-token configuration is an additional capacity control.}
\label{tab:mme_budget}
\begin{tabular}{@{}lrrr@{}}
\toprule
Method & Overall & Reasoning & Perception \\
\midrule
LVR              & 49.45 & 43.07 & 53.55 \\
GAP (36 tokens)  & 47.77 & 43.07 & 50.79 \\
GAP (144 tokens) & 50.16 & 45.20 & 53.35 \\
\bottomrule
\end{tabular}
\end{table}

\subsection{Geometry Subset and Information-Preservation Scope}
\label{app:geometry_subset}

Low-variance visual directions can contain task-relevant information, so variance retention alone does not guarantee preservation of such information.
As an additional empirical check, we evaluate the 208 geometry-problem-solving examples in MathVista using five evaluator seeds.
Table~\ref{tab:geometry_subset} shows a 5.39-point improvement in mean accuracy over the base Qwen2.5-VL-7B model.
The reported standard deviations reflect evaluator-seed variability, not independent training runs.

\begin{table}[!ht]
\centering
\small
\caption{MathVista geometry-problem-solving subset (208 examples), evaluated using five evaluator seeds. Accuracy is reported as mean $\pm$ sample standard deviation (ddof=1) in percentage points.}
\label{tab:geometry_subset}
\begin{tabular}{@{}lc@{}}
\toprule
Model & Geometry accuracy \\
\midrule
Qwen2.5-VL-7B & $70.19 \pm 0.5888$ \\
GAP           & $75.58 \pm 0.6268$ \\
\bottomrule
\end{tabular}
\end{table}

GAP thus improves the evaluated geometry subset; this check alone does not show that PCA preserves every low-variance direction or that the gain comes from PCA alone.
PCA constrains the reconstructed latent embeddings, not all transformer hidden states, and the original-image tokens remain accessible during generation.
Consequently, this evaluation measures task performance through the complete model rather than direct recovery of discarded PCA directions.

\end{document}